\documentclass[10pt,twocolumn,letterpaper]{article}

\usepackage{iccv}
\usepackage{times}
\usepackage{epsfig}
\usepackage{graphicx}
\usepackage{amsmath}
\usepackage{amssymb}
\usepackage{xcolor, colortbl}
\usepackage[numbers,sort]{natbib} 
\usepackage{multibib}
\usepackage{graphicx}
\usepackage{booktabs}
\usepackage{multirow}
\usepackage{enumitem}
\usepackage{float}

\usepackage{enumitem}
\usepackage[T1]{fontenc}

\usepackage{color}
\usepackage{tabularx}
\usepackage{tabulary}
\usepackage{hhline}
\setlist[itemize]{noitemsep, topsep=0pt}

\definecolor{Gray}{gray}{0.9}

\usepackage[pagebackref=true,breaklinks=true,colorlinks,bookmarks=false]{hyperref}

\iccvfinalcopy 

\def\iccvPaperID{6546} 
\newcommand{\COMMENT}[1]{}

\begin{document}
	
	\title{Learning Dynamics from Kinematics: \\ Estimating 2D Foot Pressure Maps from Video Frames}
	
	\author{
		Christopher Funk$^{1,3}$\thanks{Contributed equally, order chosen alphabetically}~, Savinay Nagendra$^{1 *}$, Jesse Scott$^{1 *}$, Bharadwaj Ravichandran$^{1}$, \\ John H. Challis$^{2}$, Robert T. Collins$^{1}$, Yanxi Liu$^{1}$\\
		$^{1}$School of Electrical Engineering and Computer Science. \quad
		$^{2}$Biomechanics Laboratory.\\
		The Pennsylvania State University, University Park, PA 16802 USA \\ $^{3}$Kitware, Inc\\
		{\small
			\href{mailto:christopher.funk@kitware.com}{christopher.funk@kitware.com},
			\href{mailto:sxn265@psu.edu}{sxn265@psu.edu},
			\href{mailto:jescott@cse.psu.edu}{jescott@cse.psu.edu},
			\href{mailto:bzr49@psu.edu}{bzr49@psu.edu},
		} \\
		{\small 
			\href{mailto:jhc10@psu.edu}{jhc10@psu.edu},
			\href{mailto:rcollins@cse.psu.edu}{rcollins@cse.psu.edu},
			\href{mailto:yanxi@cse.psu.edu}{yanxi@cse.psu.edu}
		}
	}
	
	\date{}

	\maketitle

	\begin{abstract}
		Pose stability analysis is the key to understanding locomotion and control of body equilibrium, with applications in numerous fields such as kinesiology, medicine, and robotics. In biomechanics, Center of Pressure (CoP) is used in studies of human postural control and gait. We propose and validate a novel approach to learn CoP from pose of a human body to aid stability analysis. More specifically, we propose an end-to-end deep learning architecture to regress foot pressure heatmaps, and hence the CoP locations, from  2D human pose derived from video. We have collected a set of long (5min +) choreographed Taiji (Tai Chi) sequences of multiple subjects with synchronized foot pressure and video data. The derived human pose data and corresponding foot pressure maps are used jointly in training a convolutional neural network with residual architecture, named PressNET. Cross-subject validation results show promising performance of PressNET, significantly outperforming the baseline method of K-Nearest Neighbors. 
		Furthermore, we demonstrate that our computation of center of pressure (CoP) from PressNET is not only significantly more accurate than those obtained from the baseline approach but also meets the expectations of corresponding lab-based measurements of stability studies in kinesiology. 
	\end{abstract}
	
	\section{Introduction}
	In the realm of health and sports, precise and quantitative digital recording and analysis of human motion provide rich content for performance characterization and training, 
	health status assessment, and diagnosis or preventive therapy of neurodegenerative syndromes.  Analysis of gait and control of balance/equilibrium has received increasing interest
	from the research community~\cite{perry1992gait, winter1991biomechanics, peterka2004dynamic} as a way to study the complex mechanisms of the  human postural system for maintaining stable pose. Stability analysis has a wide range of applications in the fields of Healthcare, Kinesiology and Robotics to understand locomotion and replicate human body movements. Understanding body dynamics, such as foot pressure, is essential to study the effects of perturbations caused by external forces and torques on the human postural system, which changes body equilibrium in static posture as well as during locomotion~\cite{winter95balancestandandwalk}. 
	
	\begin{figure}[t] \centering
		\includegraphics[width=1.0\linewidth]{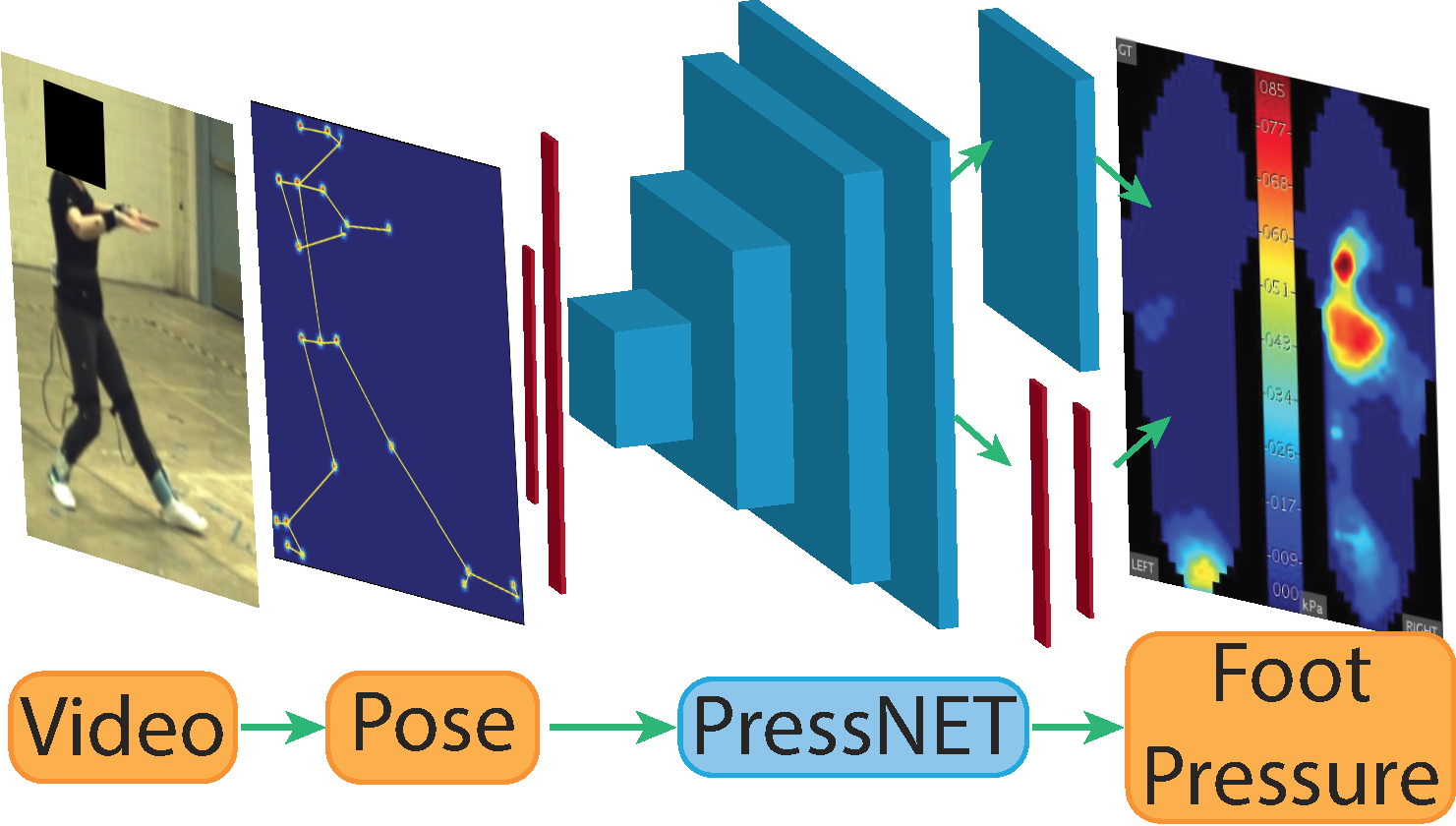}
		\vspace{-10pt}
		\caption{Our proposed PressNET network learns to predict a foot pressure heatmap from 2D human body joints extracted from a video frame using OpenPose~\cite{cao2017realtime}. The blue layers signify residual layers, the red layers signify fully connected layers.}
		\label{fig:1} \vspace{-15pt}
	\end{figure}
	
	We have chosen 24-form simplified {\em Taiji Quan}~\cite{wang_BMC_CAM2010} as a testbed for validating our computer vision and machine learning algorithms. Taiji was selected because it is low-cost, hands-free, and slow-motion sequence while containing complex body poses and movements. Taiji is practiced worldwide by millions of people of all genders, races, and ages.  Each routine lasts about 5 minutes and consists of controlled choreographed movements where the subject attempts to remain balanced and stable at all times.
	
\begin{figure}[!b] \centering
		\vspace{-10pt}
		\includegraphics[width=\linewidth]{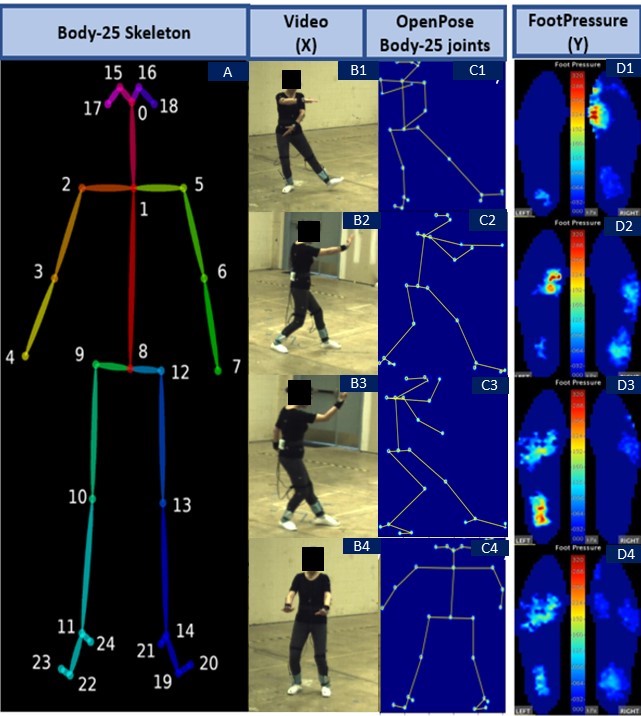}
		\vspace{-10pt}
		\caption{Column 1: \textbf{(A)}, Body25 joint set labeled by Openpose~\cite{cao2017realtime}. Column 2: \textbf{(B1 to B4)}, Video data. Column 3: \textbf{(C1 to C4)}, Corresponding Openpose detections showing the detected skeleton joints used as input to PressNET. Column 4: \textbf{(D1 to D4)}, The corresponding measured left and right foot pressure maps.}
		\label{fig:2} 
	\end{figure}

	We explore an end-to-end deep learning approach called PressNET (Figure~\ref{fig:1}) to transform kinematics (body pose) to dynamics (foot pressure), and to obtain Center of Pressure (CoP) locations from the regressed foot pressure. In order to achieve this goal, we have created the largest human motion sequence dataset of synchronized video and foot pressure data, with a total of over 700k frames (Figure \ref{fig:2}).  We represent foot pressure by an intensity heatmap that provides the distribution of pressure applied by different points of the foot against the ground, measured in kilopascals (kPa) over discretized foot sole locations. Body pose is represented by 2D human joint locations extracted using the Openpose~\cite{cao2017realtime} Body25 model on the video frames. We record video and foot pressure maps simultaneously so that there is a foot pressure map for both feet corresponding to each video frame. 
	
	Current computer vision research focuses mainly on extracting skeletal kinematics from videos, using body pose estimation and tracking to infer pose in each frame as well as the movement of body and limbs over time~\cite{cao2017realtime,bulat2016human,newell2016stacked,chen2016synthesizing,toshev2014deeppose,chen2014articulated,fan2015combining,guler2018densepose}. However, little is known whether quantitative information about dynamics can be inferred from single-view  video. While body joints and their degrees of freedom constrain the types of motion, it is the properties and actions of the muscles and weight distributions, i.e. body dynamics, that dictate the range of motion and speed produced with these degrees of freedom. Consideration of human body dynamics has been successful in explaining performance in athletics, for example the triple jump~\cite{Allen2011} and vertical jump~\cite{DomireChallis2015}. Similarly, analysis of dynamics has been used to show that strength is the limiting factor in the ability of the elderly to rise from a chair~\cite{Hughes1996}, and to determine the causes of stiff-knee gait in subjects with cerebral palsy~\cite{Goldberg2003}. An effective analysis of human movement must take into account the dynamics of the human body, and we seek an answer to the question: Can human motion dynamics be inferred from video sensors that are incapable of observing muscle activations, physical loads, and external forces directly?
	
	In biomechanics, Center of Pressure (CoP), also called Zero Moment Point (ZMP), is the point of application of the ground reaction force vector at which the moment generated due to gravity and inertia equals zero. Analysis of CoP is common in studies on human postural control and gait. Previous studies have shown that foot pressure patterns can be used to discriminate between walking subjects~\cite{pataky_etal2012, rodriguez_etal2013}. Instability of the CoP of a standing person is an indication of postural sway and thus a measure of a person's ability to maintain balance~\cite{pai2003movement, hof2008extrapolated, hof2007equations, KoEtal2015}. Knowledge of CoP trajectory during stance can elucidate possible foot pathology, provide comparative effectiveness of foot orthotics, and allow for appropriate calculation of balance control and joint kinetics during gait. CoP is usually measured directly by force plates or insole foot pressure sensors.  

	We present a method to predict foot pressure heatmaps directly from video. The major contributions and novelty of this paper are: \textbf{1) Data}: Creating the largest  synchronized video and foot pressure dataset ever recorded of a long complex human movement sequence. \textbf{2) Method}: Presenting a novel deep convolutional residual architecture, PressNET, which is the first vision-based network to regress human dynamics (foot pressure) from kinematics (body pose). \textbf{3) Application}: This is the first work seeking to compute CoP locations from video, yielding a key component for analysis of human postural control and gait stability with applications in multiple fields such as kinesiology, biomechanics, healthcare, and robotics.
	
	\section{Related Work}
	After the introduction of Deep Pose by Toshev~\etal~\cite{toshev2014deeppose}, there was a paradigm shift in the field of human pose estimation from classical approaches to deep networks. The idea of using heatmaps for ground truth data and visualization in a human pose regression problem was introduced by Tompson~\etal~\cite{tomp}, who also combine convolution layers jointly with a graphical model to represent and learn spatial relationships between joints. Many architectures use a network based on Tompson's approach~\cite{cao2017realtime, bulat2016human, newell2016stacked, chen2016synthesizing, toshev2014deeppose, chen2014articulated, fan2015combining, guler2018densepose}.
	Stacked hourglass networks by Newell~\etal~\cite{newell2016stacked} compute pose estimates using heat map regression with repeated bottom-up, top-down inferencing. An hourglass network, before stacking, is also similar to an encoder-decoder architecture, where skip connections help in preserving spatial coherence at each scale~\cite{gilbert2018fusing}, and Encoder-Decoder architectures have been extensively used for human pose estimation. Having deep residual/skip connections to preserve spatial information across multiple resolutions through the network is essential for unsupervised/semi-supervised feature learning~\cite{huang2017densely} and is a principle extensively used by densely connected convolutional networks with feed forward connections between convolution layers.  
	
	Success in 2D human pose estimation has encouraged researchers to detect 3D skeletons from image/video by extending existing 2D human pose detectors~\cite{bogo2016keep, simo2012single, chen20173d, moreno20173d, nie2017monocular, martinez2017simple} or by directly using image features~\cite{agarwal20043d, pavlakos2017coarse, zhou2016sparseness, sun2017compositional, DBLP:journals/corr/abs-1803-00455}. State-of-the-art methods for 3D human pose estimation from 2D images have concentrated on deep systems. Tome~\etal~\cite{DBLP:journals/corr/TomeRA17} proposed an estimator that reasons about 2D and 3D estimation to improve both tasks.  Zhou~\etal~\cite{zhou2017towards} augmented a 2D estimator with a 3D depth regression sub-network. Martinez~\etal~\cite{martinez2017simple} showed that given high-quality 2D joint information, the process of lifting 2D pose to 3D pose can be done efficiently using a relatively simple deep feed-forward network.
	
	All the papers discussed above concentrate on pose estimation by learning to infer joint angles or joint locations, which can be broadly classified as learning basic kinematics of a body skeleton. These methods do not delve into the external torques/forces exerted by the environment, balance, or physical interaction of the body with the scene.
	
	There have been many studies on human gait analysis~\cite{fp1, eckardt2018healthy, arvin2018step, liu2002gait} using qualitative approaches. Grimm~\etal~\cite{fp2} predict the pose of a patient using foot pressure mats.  Liu~\etal~\cite{liu2002gait} used frieze patterns to analyze gait sequences.  Although these are some insightful ways to analyze gait stability, there has been no deep learning approach to tackle this problem.  In \cite{Cai_2018}, a depth regularization model is trained to estimate dynamics of hand movement from 2D joints obtained from RGB video cameras. Papers \cite{Prevost_2013, Bacher_2014, Prevost_2016} focus on stability analysis of 3D printed models.  In this paper, we aim to use a body's kinematics to predict its dynamics and hence develop a quantitative method to analyze human stability using foot pressure derived from video.

	
	\section{Our Approach}
	
	\COMMENT{We perform a leave-one-out cross subject evaluation over the data of six subjects using our network. K-Nearest Neighbors (KNN) is applied to establish a baseline for each of the six cross-subject data splits. Mean Absolute Errors in kilopascals and RMS errors in calculated CoP locations are used as quantitative parameters to evaluate the performance of the networks. 
	}
	
	\subsection{Data Collection}
	We present the first tri-modal choreographed 24-Form Taiji sequences data set of synchronized video, motion capture,and foot pressure data (Table~\ref{tab:data_stat}). The subjects wear motion capture markers and insole foot pressure measurement sensors while being recorded. Foot pressure sensor arrays, connected to the Tekscan F-scan measurement system, are inserted as insoles in the shoes of the subject during the performance. Vicon Nexus software is used to spatiotemporally record motion capture and video data in hardware while Tekscan F-scan software is used to simultaniously record foot pressure sensor measurements that are then synchronized to the other data post collection. Motion capture data is not used in any of the experiments in this paper because:
	\begin{enumerate}[noitemsep,topsep=0pt] \setlist{nosep}
		\item We intend to create an end-to-end system to regress foot pressure maps, and hence Center of Pressure locations, directly from video;
		\item Video data collection is inexpensive and has very few hardware requirements as compared to the cumbersome process of motion capture data collection and processing; and 
		\item There are multiple existing pose prediction networks that can be used to extract 2D human body keypoints directly from video, to use as input to our network.
	\end{enumerate}
	
	\subsubsection{Video and Pose Extraction}
	Raw video data is collected at 50 fps and processed using Vicon Nexus and FFmpeg to transcode to a compressed video, with each video having its own spatiotemporal calibration. Human pose predictions are extracted from the compressed video using OpenPose~\cite{cao2017realtime}. OpenPose Body25 model uses non-parametric representations called Part Affinity Fields to regress joint positions and body segment connections between the joints. 
	The output from OpenPose thus has 3 channels, (X, Y, confidence), denoting the X and Y pixel coordinates and confidence of prediction for each of the 25 joints, making it an array of size $(25 \times 3)$.
	
	\par Figure~\ref{fig:2} (A) shows the Body25 joints labeled by OpenPose. The 25 keypoints are \textit{\{0:Nose, 1:Neck, 2:RShoulder, 3:RElbow, 4:RWrist, 5:LShoulder, 6:LElbow, 7:LWrist, 8:MidHip, 9:RHip, 10:RKnee, 11:RAnkle, 12:LHip, 13:LKnee, 14:LAnkle, 15:REye, 16:LEye, 17:REar, 18:LEar, 19:LBigToe, 20:LSmallToe, 21:LHeel, 22:RBigToe, 23:RSmallToe, 24:RHeel\}}. Figure~\ref{fig:2} (C and D) portrays sample input-output pairs used to train our network. The video frames of a subject performing 24-form Taiji, shown in Figure~\ref{fig:2} (B1 to B4), are processed through the OpenPose network to extract 25 body joint locations with respect to a fixed coordinate axis. 
	Figure~\ref{fig:2} (C1 to C4) shows the joints extracted from OpenPose.
	For training PressNET, we directly use arrays of 2D joint locations as inputs. The confidence of keypoints 15 to 18 (Eyes and Ears) are zero for more than 50\% of the video frames due to occlusions. 
	
	\begin{table}[!t] \centering
		\resizebox{1.00\linewidth}{!}{%
			\begin{tabular}{*{8}{c}} \hline
				Subject & Session & Take & \# frames & Mean & Std & Median & Max \\ \hline
				\multirow{9}{*}{1} & \multirow{3}{*}{1} & 1 & 17995 & 3.73 & 11.14 & 3.67 &  282 \\ \cline{3-8}
				&                    & 2 & 17625 & 3.48 & 10.61 & 3.67 &  299 \\ \cline{3-8}
				&                    & 3 & 17705 & 3.29 & 10.34 & 3.21 &  219 \\ \cline{2-8}
				& \multirow{3}{*}{2} & 1 & 17705 & 6.18 & 20.21 & 6.01 &  417 \\ \cline{3-8}
				&                    & 2 & 17355 & 5.75 & 19.18 & 5.55 &  476 \\ \cline{3-8}
				&                    & 3 & 17625 & 5.41 & 18.56 & 5.23 &  521 \\ \cline{2-8}
				& \multirow{3}{*}{3} & 1 & 17580 & 5.40 & 19.69 & 5.26 &  617 \\ \cline{3-8}
				&                    & 2 & 17645 & 5.34 & 19.39 & 5.21 &  636 \\ \cline{3-8}
				&                    & 3 & 17685 & 5.35 & 19.47 & 5.15 &  582 \\ \hline
				\multirow{11}{*}{2} & \multirow{3}{*}{1} & 1 & 13230 & 7.80 & 37.62 & 7.72 & 1000 \\ \cline{3-8}
				&                    & 2 & 13335 & 6.71 & 35.20 & 6.68 & 1000 \\ \cline{3-8}
				&                    & 3 &  9500 & 6.13 & 33.93 & 6.03 & 1000 \\ \cline{2-8}
				& \multirow{4}{*}{2} & 1 & 14105 & 5.26 & 30.18 & 5.18 & 1000 \\ \cline{3-8}
				&                    & 2 &  6475 & 4.81 & 30.64 & 4.73 & 1000 \\ \cline{3-8}
				&                    & 3 & 13885 & 4.68 & 28.62 & 4.56 & 1000 \\ \cline{3-8}
				&                    & 4 & 12185 & 3.02 & 21.05 & 2.83 & 1000 \\ \cline{2-8}
				& \multirow{4}{*}{3} & 1 &  5600 & 2.95 & 21.15 & 2.71 & 1000 \\ \cline{3-8}
				&                    & 2 &  6845 & 3.25 & 23.23 & 3.09 & 1000 \\ \cline{3-8}
				&                    & 3 & 12135 & 3.08 & 21.22 & 2.86 & 1000 \\ \cline{3-8}
				&                    & 4 &  8725 & 2.93 & 20.05 & 2.71 & 1000 \\ \hline
				\multirow{9}{*}{3} & \multirow{3}{*}{1} & 1 & 11210 & 4.41 & 17.37 & 4.45 &  614 \\ \cline{3-8}
				&                    & 2 & 10605 & 3.88 & 15.39 & 3.91 &  614 \\ \cline{3-8}
				&                    & 3 & 11075 & 3.34 & 14.23 & 3.34 &  683 \\ \cline{2-8}
				& \multirow{3}{*}{2} & 1 & 11295 & 6.17 & 28.09 & 6.15 & 1000 \\ \cline{3-8}  
				&                    & 2 & 10700 & 5.69 & 26.88 & 5.67 & 1000 \\ \cline{3-8}
				&                    & 3 & 10945 & 5.19 & 25.53 & 5.26 & 1000 \\ \cline{2-8}
				& \multirow{3}{*}{3} & 1 & 12410 & 5.80 & 30.57 & 5.59 & 1000 \\ \cline{3-8}
				&                    & 2 & 11805 & 5.31 & 28.58 & 5.18 & 1000 \\ \cline{3-8}
				&                    & 3 & 11950 & 5.49 & 28.95 & 5.48 & 1000 \\ \hline
				\multirow{11}{*}{4} & \multirow{3}{*}{1} & 1 & 13115 & 7.37 & 24.76 & 7.40 &  679 \\ \cline{3-8}
				&                    & 2 & 13715 & 6.05 & 21.75 & 6.10 &  775 \\ \cline{3-8}
				&                    & 3 & 13015 & 5.25 & 20.07 & 5.27 &  650 \\ \cline{2-8}
				& \multirow{3}{*}{2} & 1 & 15405 & 8.12 & 31.11 & 8.24 & 1000 \\ \cline{3-8}
				&                    & 2 & 14370 & 7.62 & 30.11 & 7.68 & 1000 \\ \cline{3-8}
				&                    & 3 &  9370 & 6.05 & 25.72 & 6.05 &  781 \\ \cline{2-8}
				& \multirow{5}{*}{3} & 1 & 14340 & 8.05 & 35.20 & 8.05 & 1000 \\ \cline{3-8}
				&                    & 2 & 13685 & 7.60 & 32.50 & 7.66 & 1000 \\ \cline{3-8}
				&                    & 3 & 13675 & 7.45 & 32.15 & 7.57 & 1000 \\ \cline{3-8}
				&                    & 4 & 13015 & 7.38 & 32.69 & 7.46 & 1000 \\ \cline{3-8}
				&                    & 5 & 13045 & 7.27 & 31.52 & 7.26 & 1000 \\ \hline
				\multirow{8}{*}{5} & \multirow{4}{*}{1} & 1 & 18000 & 6.85 & 25.81 & 6.95 & 1000 \\ \cline{3-8}
				&                    & 2 & 17300 & 6.83 & 25.81 & 6.90 & 1000 \\ \cline{3-8}
				&                    & 3 & 18005 & 7.49 & 28.11 & 7.58 & 1000 \\ \cline{3-8}
				&                    & 4 & 16750 & 8.18 & 30.21 & 8.26 & 1000 \\ \cline{2-8}
				& \multirow{4}{*}{2} & 1 & 16545 & 8.06 & 32.01 & 7.84 & 1000 \\ \cline{3-8}
				&                    & 2 &  4000 & 8.39 & 32.42 & 8.40 & 1000 \\ \cline{3-8}
				&                    & 3 & 16910 & 8.94 & 34.47 & 8.60 & 1000 \\ \cline{3-8}
				&                    & 4 & 16440 & 9.63 & 37.12 & 9.34 & 1000 \\ \hline
				\multirow{10}{*}{6} & \multirow{6}{*}{1} & 1 & 17395 & 6.79 & 30.20 & 6.78 &  887 \\ \cline{3-8}
				&                    & 2 & 16330 & 6.87 & 30.43 & 6.87 & 1000 \\ \cline{3-8}
				&                    & 3 & 15760 & 7.04 & 31.26 & 7.01 & 1000 \\ \cline{3-8}
				&                    & 4 & 15575 & 7.23 & 31.93 & 7.20 & 1000 \\ \cline{3-8}
				&                    & 5 & 15810 & 7.46 & 32.90 & 7.44 &  983 \\ \cline{3-8}
				&                    & 6 & 16095 & 7.61 & 33.54 & 7.60 & 1000 \\ \cline{2-8}
				& \multirow{4}{*}{2} & 1 & 15520 & 6.39 & 27.98 & 6.30 &  764 \\ \cline{3-8}
				&                    & 2 & 15000 & 6.69 & 29.88 & 6.53 &  951 \\ \cline{3-8}
				&                    & 3 & 15200 & 6.87 & 30.38 & 6.70 &  979 \\ \cline{3-8}
				&                    & 4 & 15140 & 7.13 & 31.43 & 7.07 &  924 \\ \hline
			\end{tabular}%
		}
		\vspace{1pt}
		\caption{Foot pressure data statistics showing number of frames, mean, standard deviation, median, and maximum intensity of foot pressure in kilopascals (kPa) per take and session for each subject. As a results, we have a total of 794,035 video-foot pressure frame pairs for training-testing.
			\label{tab:data_stat}} \vspace{-15pt}
	\end{table}
	
	\subsubsection{Foot Pressure}\label{fp}
	Foot pressure is collected at 100 fps using a Tekscan F-Scan insole pressure measurement system (Figure~\ref{fig:2} right column). Each subject is provided a pair of canvas shoes outfitted with cut-to-fit capacitive pressure measurement insoles. 
	Sensor noise floor is approximately 3 KPa. 
	%
	The foot pressure heatmaps generated are 2-channel images of size $60 \times 21 $ as shown in Figure~\ref{fig:2} (D1 to D4),  
	%
	%
	and synched with the mocap data (and thus to the video).
	
	\COMMENT{the derivative of average foot pressure divided by 
		In post processing, the foot pressure data To synchronize the foot pressure data, Each frame has a mean calculated by taking the 60x21 pressure down to a single value. This value is based on the active sensor area at each time sample that is then filtered with a 1 second averaging filter. Then, the derivative of the time sequence is subtracted by the mean and divided by the standard deviation of foot pressure. The MoCap is the mean position of the ankle, toe and heal markers. This position is also filtered with a 1 second averaging filter and then the derivative of the result is taken and is normalized by subtracting the mean and dividing by the standard deviation. MoCap is related to video (50 Hz) as every other frame. MoCap and video are synced via Vicon's Nexus.}
	
	\begin{table}[!t] \centering
		\resizebox{0.85\linewidth}{!}{
			\begin{tabular}{*{4}{c}} \hline
				Subject & Training Set & Validation Set & Test Set \\ \hline
				1 &      569,815 &         65,300 &  158,920 \\
				2 &      603,755 &         74,260 &  116,020 \\
				3 &      622,835 &         69,205 &  101,995 \\
				4 &      582,555 &         64,730 &  146,750 \\
				5 &      603,075 &         67,010 &  123,950 \\
				6 &      572,590 &         63,620 &  157,825 \\ \hline
			\end{tabular}
		}
		\vspace{1pt}
		\caption{The number of training, validation, and test frames of each train-test split for the leave one subject out segmentation of the 794,035 frame dataset.}
		\label{tab:data_splits} \vspace{-10pt}
	\end{table}
	
	\begin{table}[!t] \centering
		\resizebox{1.00\linewidth}{!}{%
			\begin{tabular}{*{5}{c}} \hline
				Subject & Total Frames & Weight (kg) & Height (m) & Gender \\ \hline
				1 &     158,920 &          52 &       1.60 & Female \\
				2 &     116,020 &          67 &       1.72 &   Male \\
				3 &     101,995 &          64 &       1.60 & Female \\
				4 &     146,750 &          77 &       1.70 &   Male \\ 
				5 &     123,950 &          60 &       1.56 & Female \\
				6 &     157,825 &          55 &       1.54 & Female \\ \hline
			\end{tabular}%
		}
		\vspace{1pt}
		\caption{Dataset statistics showing demographic information including subject weight(kg), height(mm) and gender. We have 3 professionals (20+ years experience) and 3 amateurs (5-10 years experience).
			\label{tab:subject_stats} \vspace{-5pt}
		}
	\end{table}
	
	
	\begin{figure*}[!t] \centering
		\resizebox{1\linewidth}{!}{
			\setlength{\tabcolsep}{-25pt}
			\begin{tabular}{cccc}

				\includegraphics{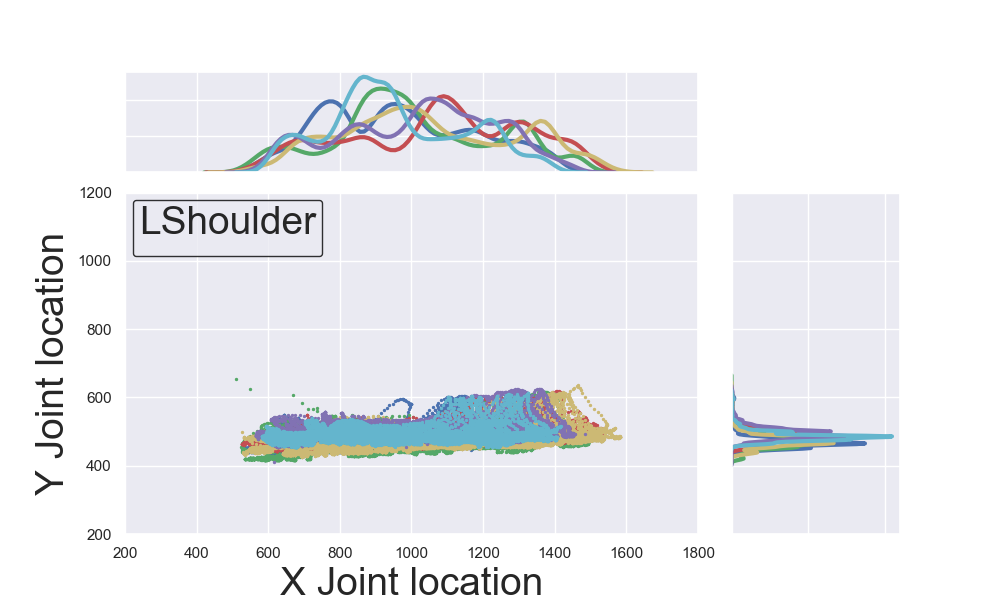} & \includegraphics{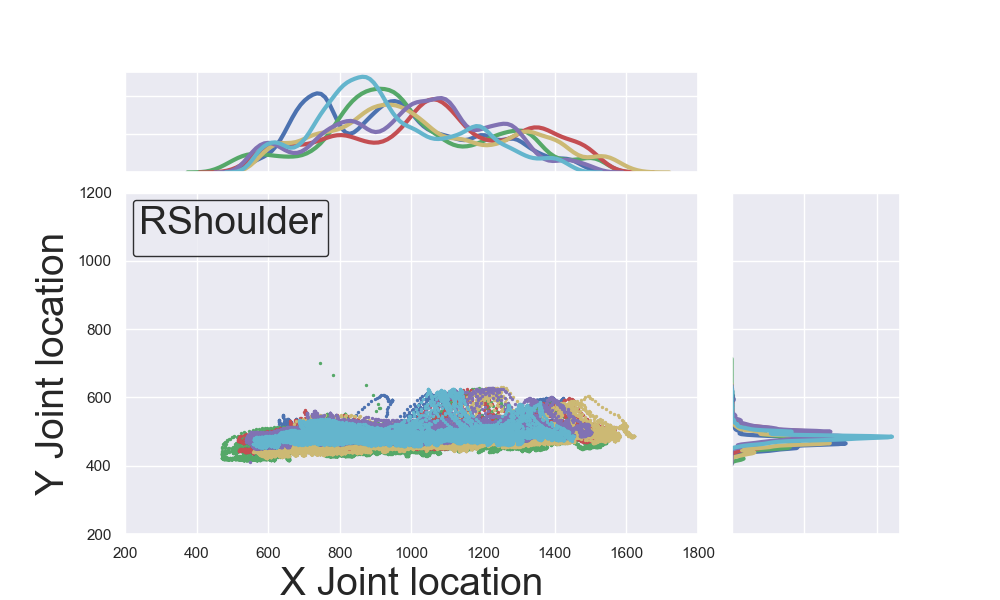} & 
				\includegraphics{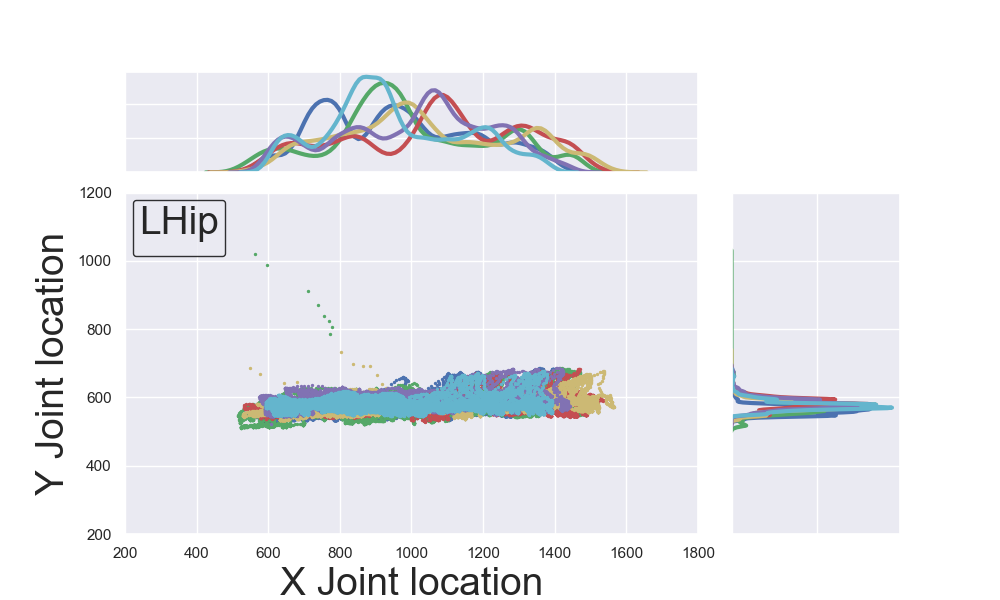}      & \includegraphics{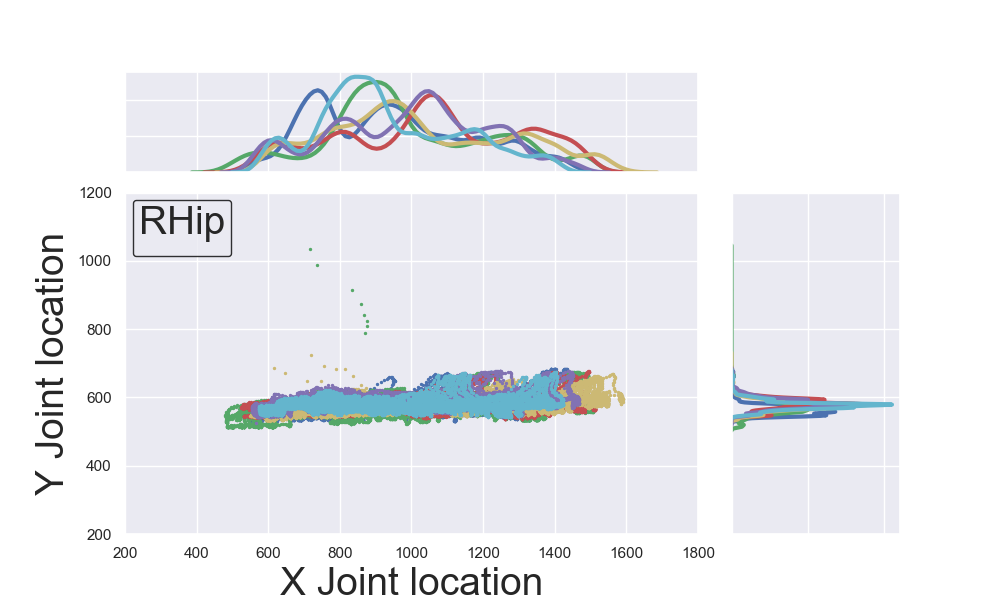}      \\      
				\includegraphics{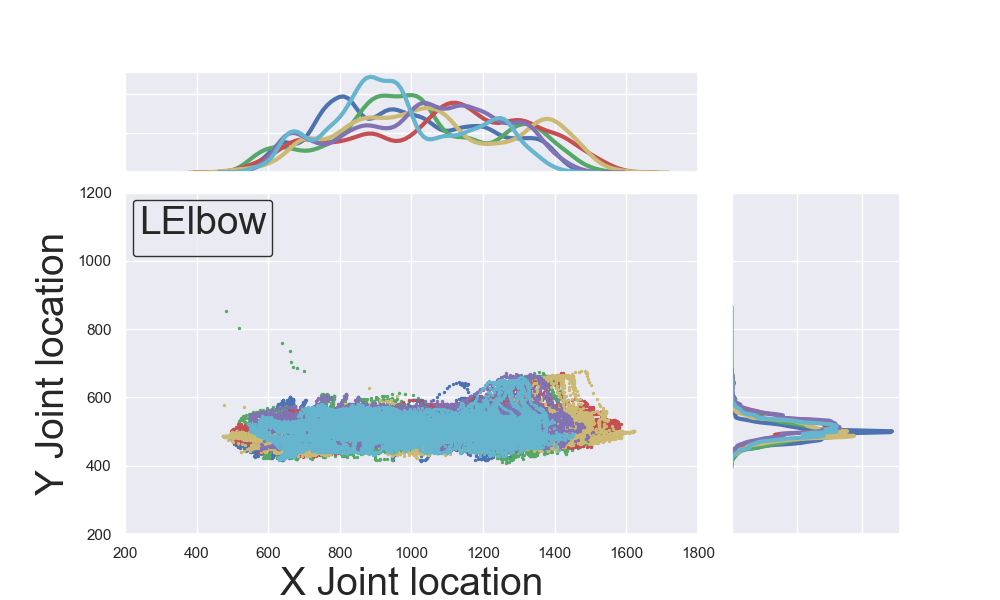}    & \includegraphics{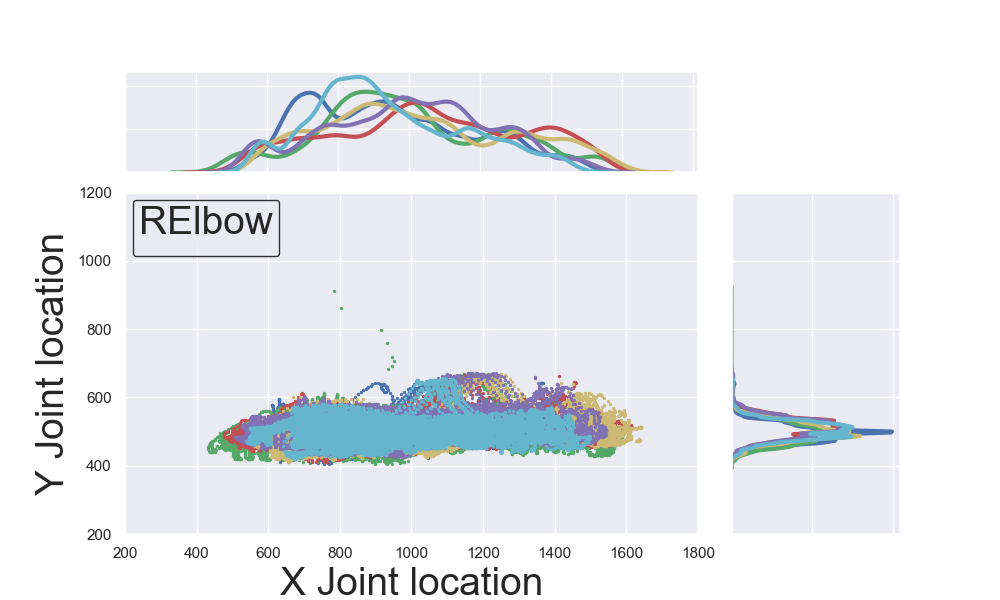}    & 
				\includegraphics{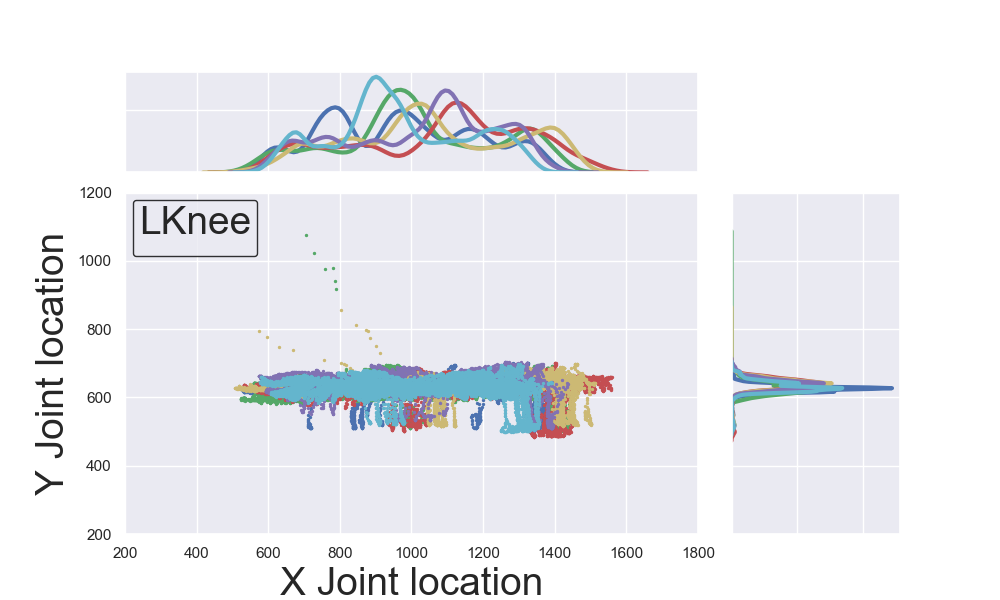}     & \includegraphics{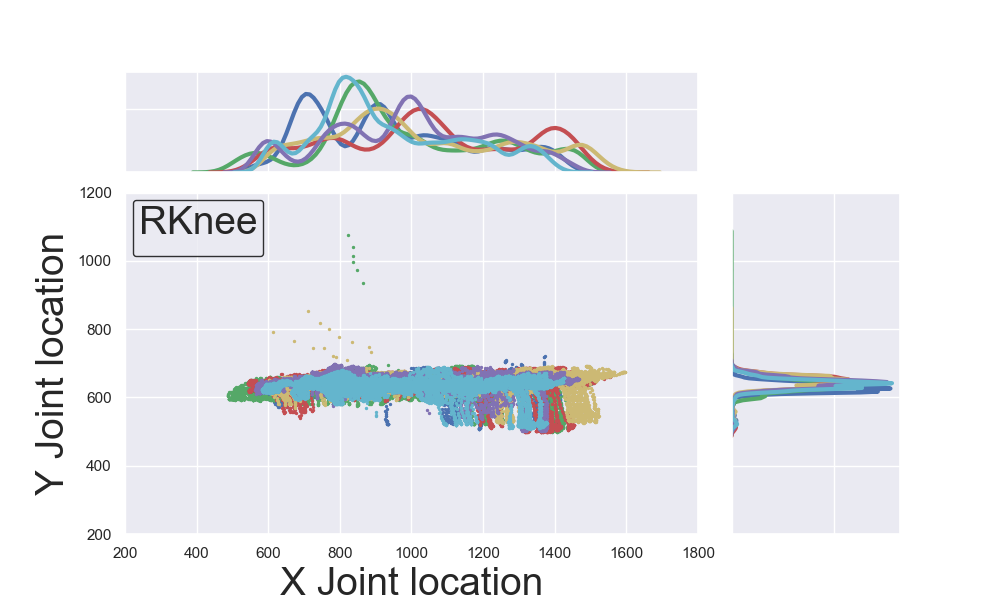}     \\
				\includegraphics{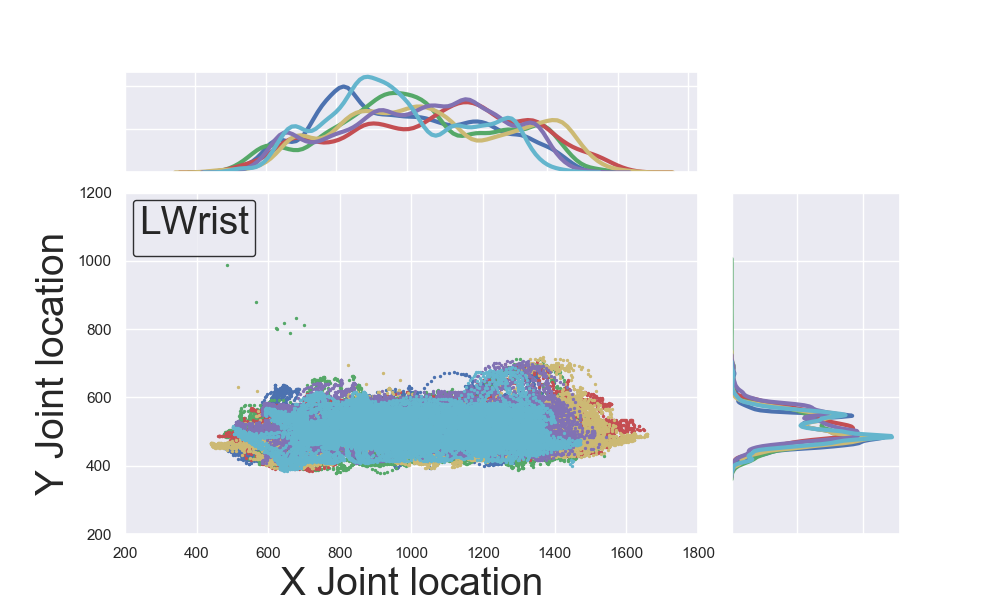}    & \includegraphics{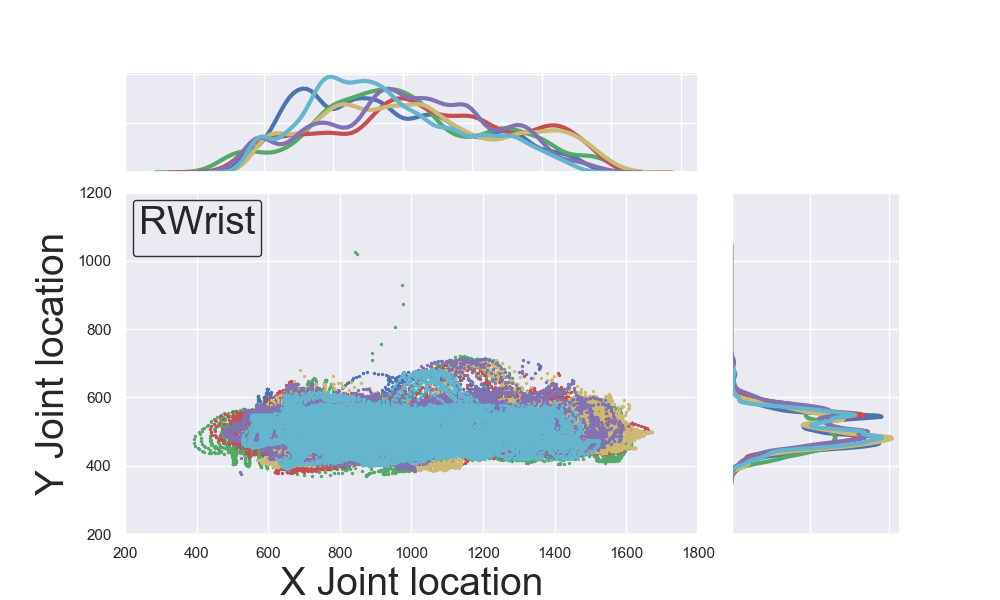}    &
				\includegraphics{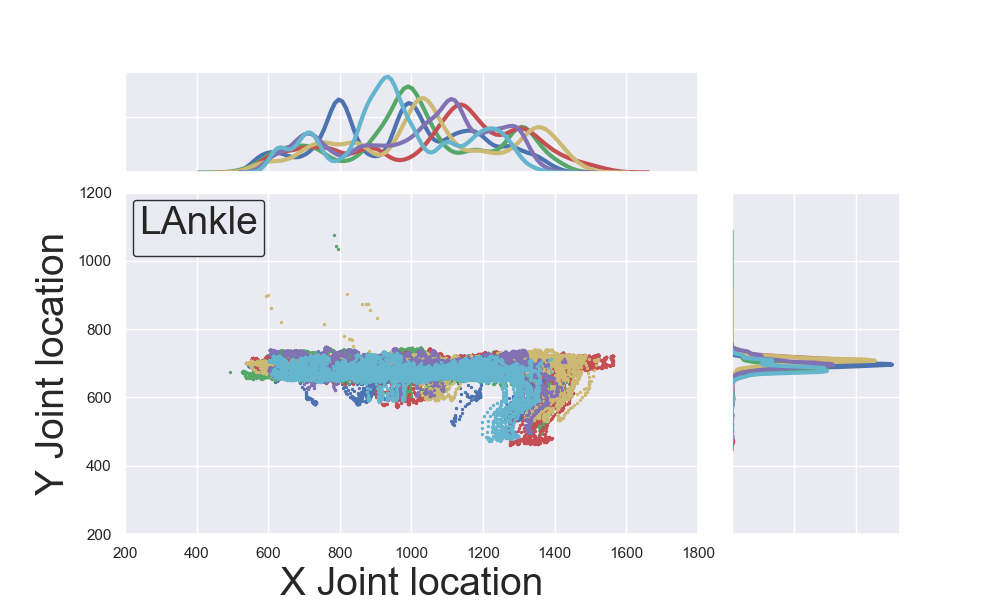}    & \includegraphics{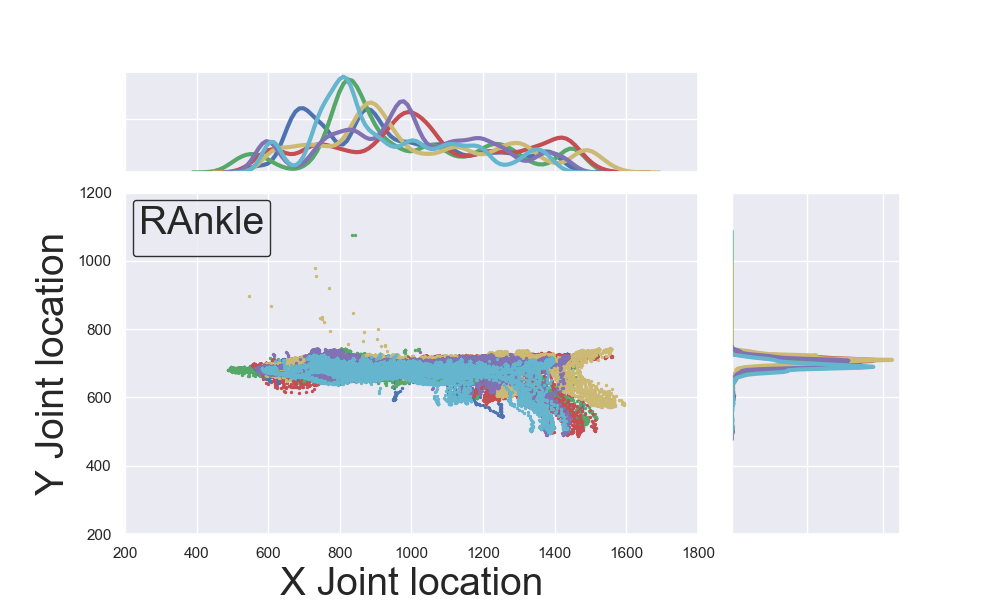}    \\
				\includegraphics{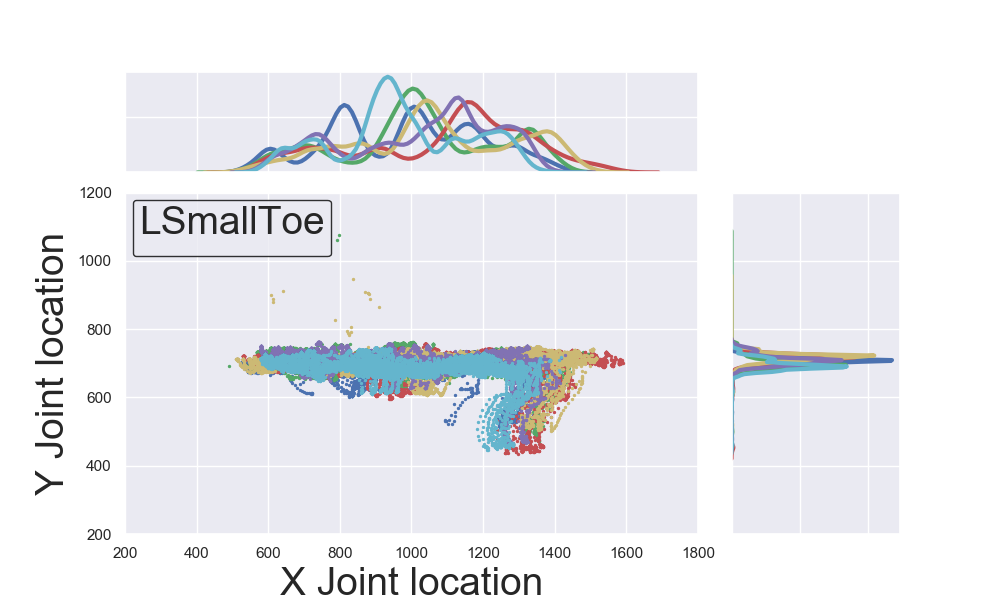} & \includegraphics{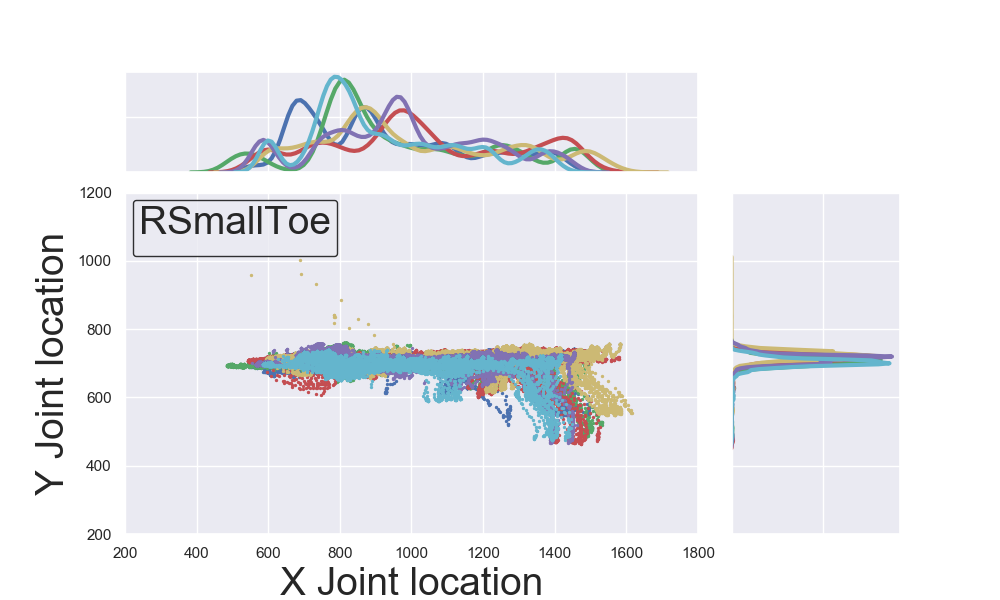} &
				\includegraphics{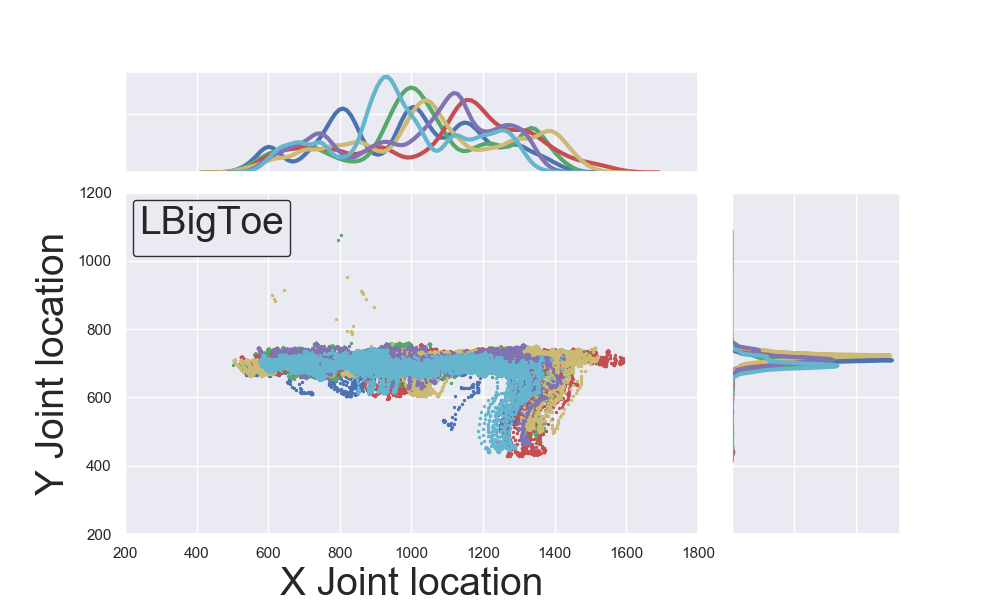}   & \includegraphics{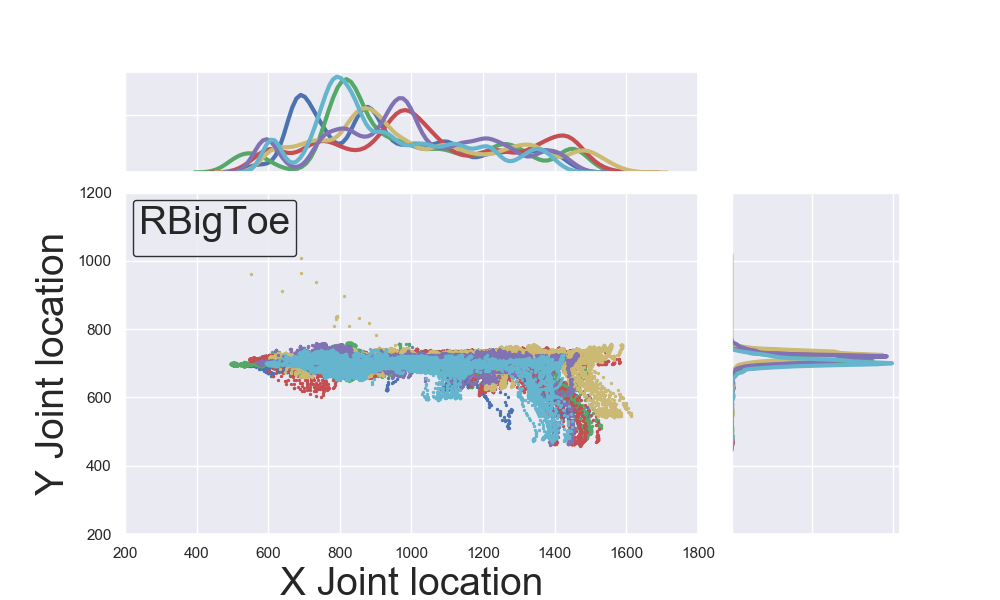}   \\
				\includegraphics{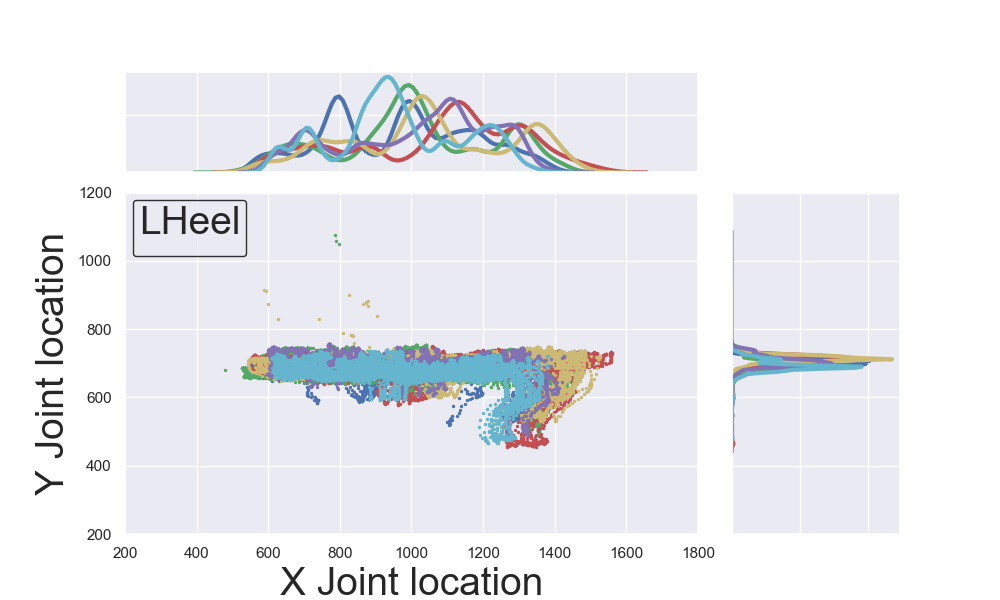}     & \includegraphics{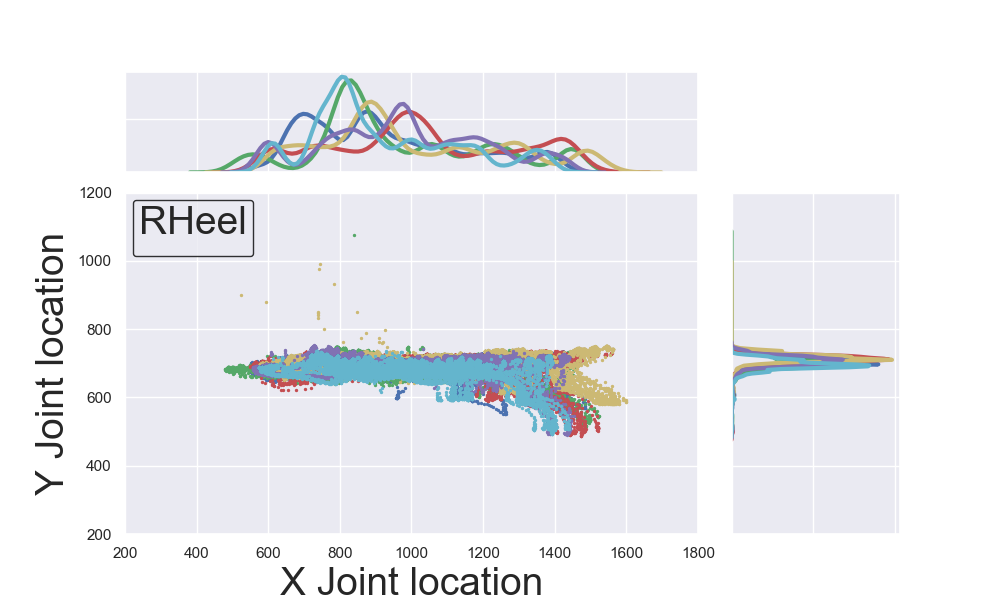}     &
				\includegraphics{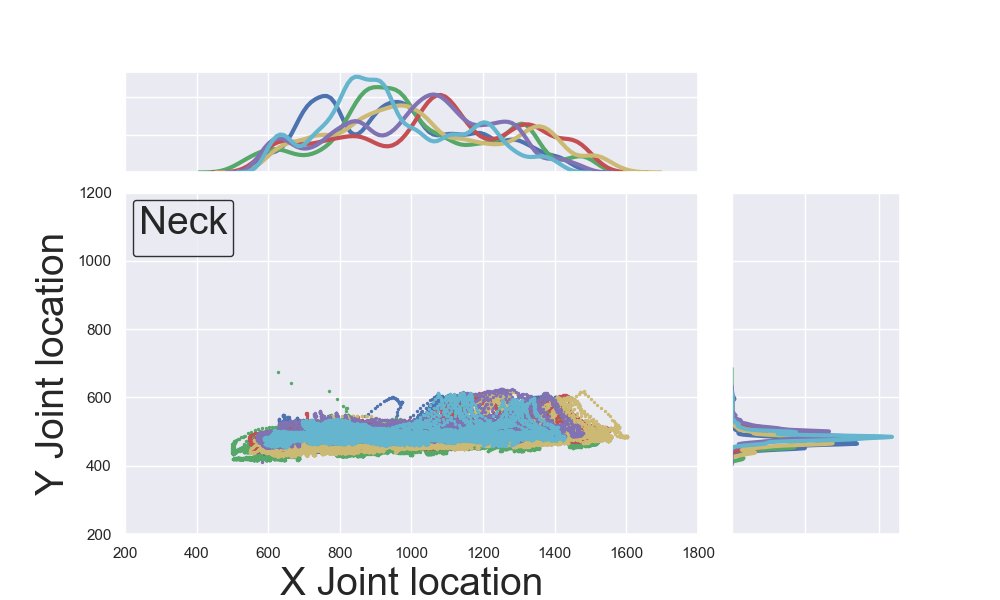}      & \includegraphics{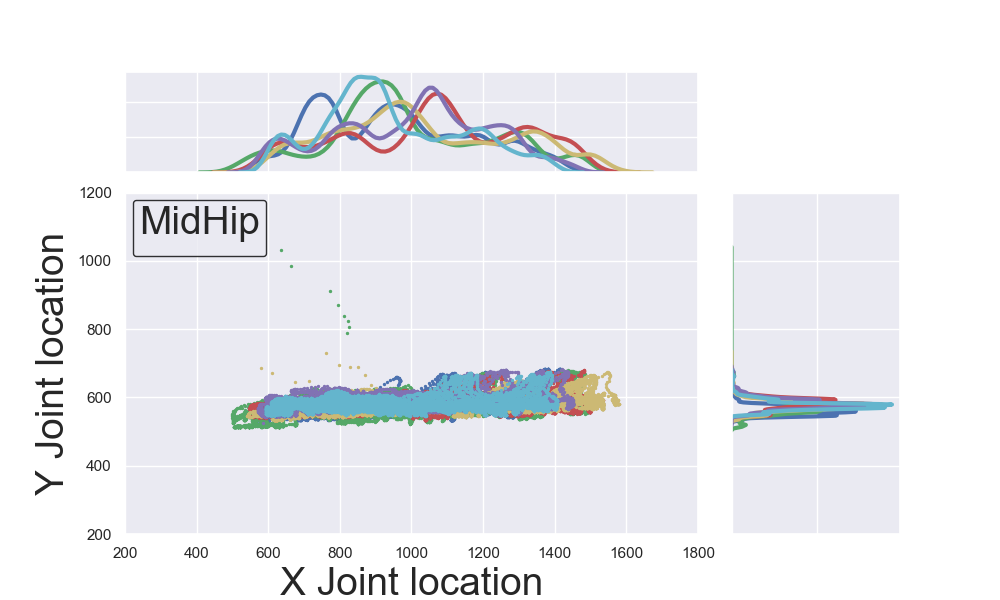}    \\
			\end{tabular}
		}
		\vspace{1pt}
		\caption{Per body joint Kernel Density plots and 2D scatter plots of OpenPose~\cite{cao2017realtime} data for all subjects. The datapoints for different subjects are represented with different colors. (Sub1 - Blue, Sub2 - Orange, Sub3 - Green, Sub4 - Red, Sub5 - Purple, Sub6 - Brown)}
		\label{fig:hist}  \vspace{-10pt}
	\end{figure*}
	
	\definecolor{myyellow}{RGB}{254, 254, 8}
	
	\begin{figure*}[!t] \centering
		\resizebox{1.00\linewidth}{!}{
			\includegraphics{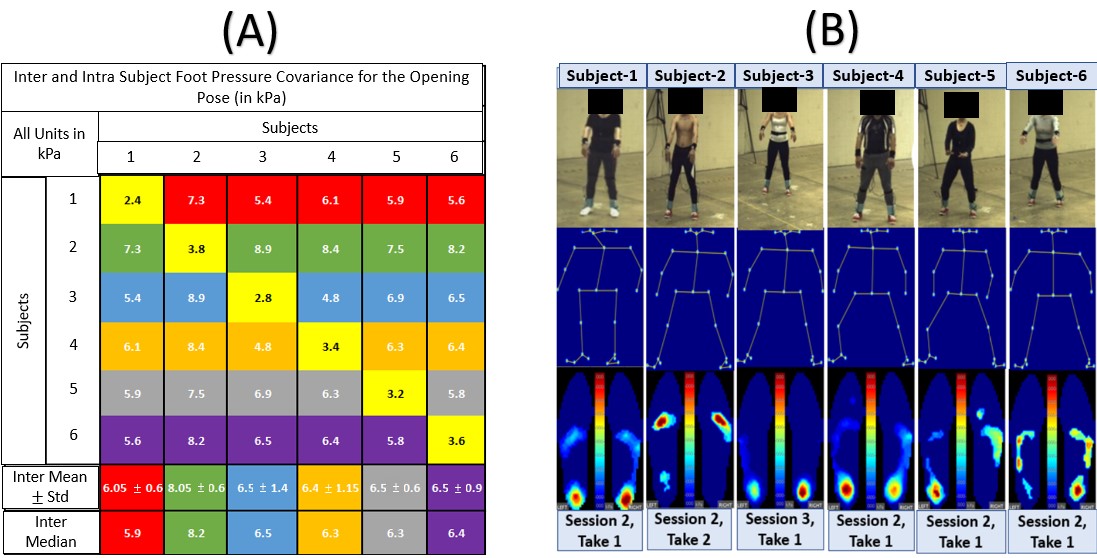}
		}
		\vspace{-10pt}
		\caption{\textbf{(A)}: Inter- and intra(yellow)- subject foot pressure mean absolute difference for the opening pose of 24-form Taiji. Rowwise colors (Red, Green, Blue, Orange, Grey and Purple) represent information of Subjects 1, 2, 3, 4, 5, and 6 respectively \textbf{(B)}: Opening pose (video), extracted skeleton and corresponding foot pressure of the six subjects.}
		\label{data_stat} \vspace{-5pt}
	\end{figure*}
	
	\subsubsection{Data Statistics}
	
	To justify the adequacy of our data set for a machine learning task, we make few initial data statistics observations. Ultimately, our leave-one-subject-out cross validation experimental results provide a quantified validation of our method and dataset used.

	Table~\ref{tab:data_stat} provides complete information about the foot pressure dataset. 
	A "take" refers to >5 min long continuous motion sequences. A "session" refers to a collection of "takes".
	Each subject performs 2 to 3 sessions of 24-form Taiji at an average of 3 takes per session, amounting to a total of 794,035 frames of video-foot pressure paired data. 
	We have observed that:
	\begin{itemize}
		\item (1) {\bf Subject demographics}: there is sufficient demographics diversity in the  subjects in terms of their gender, age, weight, height and years of experience in Taiji practice and  professional levels (Table~\ref{tab:subject_stats}). The range of experience in Taiji of our subjects varies from amateurs (5-10 years) to professionals (20+ years). We currently have 3 amateurs and 3 professionals in our dataset.
		%
		\item (2) {\bf Body joint} (from video) value statistical distributions: Figure~\ref{fig:hist} shows per feature kernel density plots of joints extracted from the OpenPose network.  These distributions support the hypothesis that the subjects are statistically similar.   

		%
		\begin{figure*}[!t] \centering
			\includegraphics[width=.96\linewidth]{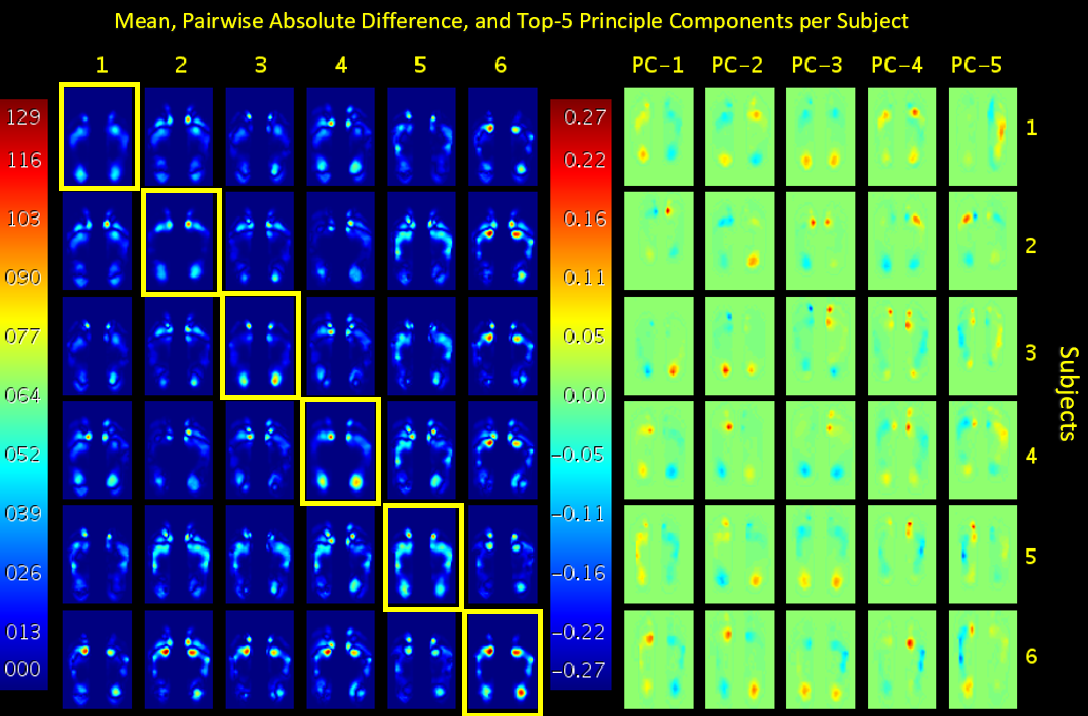}
			\caption{Left: Pairwise absolute difference between the mean foot pressure across all 6 subjects. Mean pressure is provided on diagonal (yellow bounding box). Right: Top-5 Principal Components of Foot Pressure data per subject.}
			\label{fig:diffPCA}\vspace{-15pt}
		\end{figure*}
		
		%
		
		\item (3) {\bf Foot-pressure} variation under the same pose:
		In order to depict inter/intra-subject foot pressure covariance information for the opening pose of 24-form Taiji, Figure~\ref{data_stat} (A) shows a color coded table. It can be observed from Figure~\ref{data_stat} (B) that the foot pressure maps are visually different for the same pose, therefore establishing a lower bound on errors for estimating foot pressure from a single pose.  The learning system cannot do better than this inherent difference.  
		\item (4) {\bf PCA analysis}:  Figure~\ref{fig:diffPCA} highlights the inter-subject and intra-subject variance of foot pressure data via PCA analysis. The left portion of Figure~\ref{fig:diffPCA} shows the mean foot pressure for each individual subject on the diagonal and the difference of means for pairs of subjects off-diagonal, for all the subjects.  The difference of mean pressure highlights that each subject has a unique pressure distribution relative to other subjects. The right portion of Figure~\ref{fig:diffPCA} highlights the top-5 principal components of the foot pressure map data for each subject, with the rows represent individual subjects. From Figure~\ref{fig:diffPCA} we can see that each principal component encodes different types of information (variability in left/right foot pressure, in toe/heel pressure, and so on), and that the collection of top PCs encode similar modes of variation, although not in the exact same order (for example, Subject 1's 1st principal component encodes pressure shifts between the left and right foot, whereas, Subject 2's 2nd principal component encodes that information). 
	\end{itemize}
	
	\COMMENT{
		\par The largest publicly available dataset of motion capture data, Human 3.6M~\cite{ionescu2014human3} has a total of 15 varied activities from 11 subjects (6 male and 5 female), where 7 subjects are used for training and validation and 4 for testing. For a particular activity, the total number of frames in Human 3.6M is approximately 250,000. Our data has triple the number of frames from a total of 57 videos each depicting a single, long activity (24-Form Taiji). In the dataset, length and complexity of recorded performances are traded for number of subjects. Furthermore, a leave-one-subject-out cross-validation is performed as is commonly done when using Human 3.6M. This choice provides a moderate amount of body shape variability as well as different ranges of mobility, ensuring that the network does not overfit and learns to neglect outliers in the input data distribution. 
	}

	\subsection{Network and Training}
	The design of our network is initially motivated by the residual generator of the Improved Wasserstein GAN~\cite{gulrajani2017improved}. We use a generator-inspired architecture since our input is 1D and the output is a 2D heatmap. This design aids in capturing information at different resolutions, acting like a decoder network for feature extraction. The primary aim of our network is to extract features without loss of spatial information across different resolutions. We try to learn the correlation between the pose, encoded as 25 joint locations, and the corresponding foot pressure map intensity distribution. We train a Convolutional Residual architecture, PressNET, to regress foot pressure distributions from a given pose, over data from multiple subjects using a leave-one-subject-out strategy. We do not use any discriminator since ground truth data is available. Thus, this is formulated as a supervised learning (regression) problem.
	
	\subsubsection{Data Pre-Processing}\label{dp}
	Input body pose data from openpose is an array of size $(25\times 3)$. We use hip joint as the center point to remove camera specific offsets during video recording. The hip joint is $(0,0)$ after centering and is removed from the training and testing data sets. Data is normalized per body joint by subtracting the feature's mean and dividing by its standard deviation, leading to a zero-mean, unit variance distribution. Zero confidence (undetected) OpenPose joints are not considered during normalization. Confidence of joint detections is then removed and not used for training. After pre-processing and normalization, the input array is of size $(24\times 2)$, which is flattened to a 1D vector of size $(48\times1)$ and used as input to our network.
	
	\begin{figure*}[!t] \centering
		\includegraphics[width=0.95\linewidth]{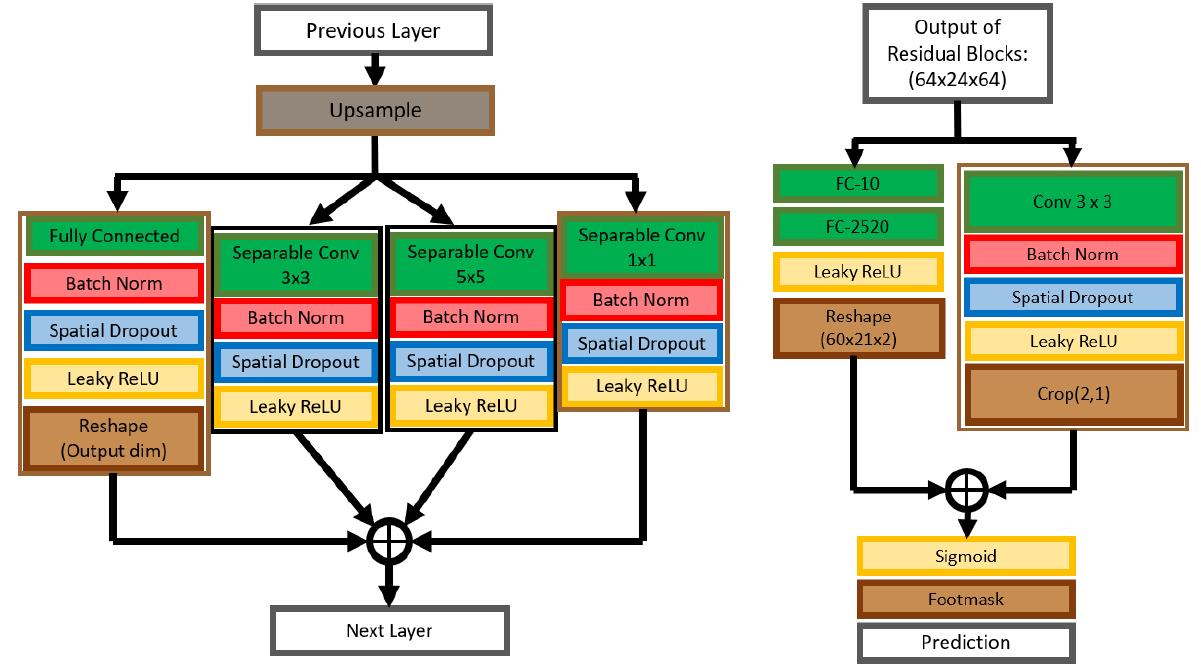}
		\vspace{-5pt}
		\caption{Left: A residual block, one of the building blocks of PressNET network, upsamples the data and computes features. Right: Final set of layers of PressNET include a fully connected layer layer and a concurrent branch to preserve spatial consistency.} 
		\label{fig:res2} \vspace{-10pt}
	\end{figure*}
	
	Foot pressure data, which is originally recorded in kilopascals (kPa), has invalid prexels marked as Not a Number ({NaN}) representing regions outside the footmask. These prexels are set to zero since the network library cannot train with NaN values. Any prexel values greater than 1000 kilopascals are clipped to 1000 to remove noise from the dataset. Data is converted from kilopascals to PSI (Pounds per Square Inch) by multiplying with a constant  $0.145$. The dimension of a single cell in the insole sensor array is $(0.508\text{cm} \times 0.508\text{cm})$, thus having an area of $0.258\text{cm}^2$ or $0.039\text{inch}^2$. The data in PSI is multiplied by this sensor area and divided by the weight of the subject reported in Table~\ref{tab:data_splits}. Thus, the foot pressure distribution is now weight-normalized, unit-less and independent of the subject. Furthermore, the data is normalized by dividing each prexel by its max intensity in the distribution. The left and right normalized foot pressure maps are concatenated as two channels of a resulting ground truth foot pressure heatmap of size $(60 \times 21 \times 2)$, with prexel intensities in the range $[0,1]$.
	
	\subsubsection{Network}
	The PressNET network is a feed forward Convolutional Neural Network which inputs a 1D vector of joints and outputs a 2D foot pressure (Figure~\ref{fig:1}). The input layer is a flattened vector of joint coordinates of size $48\times 1$ (24 joints $\times$ 2 coordinates since the mid hip joint is removed), which contains the kinematic information about a pose. The input is processed through a fully connected layer with an output dimension of $6144\times 1$. This output is reshaped into an image of size $4\times3$ with 512 channels. The network contains four residual convolution blocks that perform nearest neighbor upsampling. The first block upsamples the input by $(2,1)$ and the other three upsample by 2. 
	
	
	The residual block of PressNET, shown in Figure~\ref{fig:res2} Left, has three parallel convolution layers with kernel sizes $5\times5$ and $3\times3$ and a residual $1\times1$. There is an additional parallel fully connected layer, which takes the upsampled input and returns a flattened array of dimension equal to the output dimension of the residual block. This output is reshaped and added to the output of the other three parallel layers to constitute the output of the block.  The number of channels of each residual block is progressively halved as the resolution doubles, starting at 512 channels and decreasing to 64. 
	
	The output of the final residual block is split, flattened and sent to a convolutional branch and a fully connected branch. The convolutional branch contains a $3 \times 3 $ normal convolution layer to get a 2 channel output of shape $64\times 24$ and cropped to size of the foot pressure map ($60\times21\times2$). On the fully connected branch, the activations are run through multiple fully connected layers and then reshaped to the size of the foot pressure map. The sizes of the fully connected layers for PressNET are 10 and 2520 (Figure~\ref{fig:res2} Right).  The output of these branches are added together and then a foot pressure mask is applied to only learn the valid parts of the data. Finally, a sigmoid activation is used to compress the output into the range [0,1]. The convolutional branch serves to preserve spatial coherence while the fully connected branch has a field of view over the entire prediction. With the combined spatial coherence of the concurrent branch and fully connected layers in every residual convolutional block, PressNET has $\sim$3 million parameters.
	
	All convolutional layers are separable convolution layers that split a kernel into two to perform depth-wise and point-wise convolutions.  Separable convolutions reduce the number of network parameters as well as increase the accuracy~\cite{chollet2017xception}. Batch normalization~\cite{ioffe2015batch} and spatial dropouts~\cite{tompson2015efficient} are applied after every convolution layer. Leaky ReLU~\cite{maas2013rectifier} is used as a common activation function throughout the network, except the output layer.
	
	\subsubsection{Training Details}
	We evaluate our network on six splits of the dataset. Our dataset is split by subject in a leave-one-subject-out cross-validation. The validation data consists of the last take from each subject used in the training data. The goal of this cross-subject validation is to show how well this network generalizes to an unseen individual. PressNET is trained with a learning rate of $1e^{-3}$ for 20 epochs at a batch size of 32 for all splits on a NVIDIA Tesla P100 GPU cluster with 8GB of memory. Data pre-processing is carried out before training as mentioned in Section~\ref{dp}.
	PressNET takes 3 to 3.5 hours to train on each split. The problem is formulated as a regression with a sigmoid activation layer as the last activation layer since the output data is in the range $[0,1]$.  A binary footmask having ones marked for valid prexels and zeros marked for invalid prexels (produced by the foot pressure capturing system) is element-wise multiplied in the network. This enables the network to not have to learn the approximate shape of the foot in the course of training and solely learn foot pressure. The learning rate is reduced to $1e^{-5}$ after 12 epochs to ensure a decrease in validation loss with training. Mean Squared Error (MSE) is used as the loss function along with Adam Optimizer for supervision, as we are learning the distribution of prexels~\cite{bishop2006pattern}. 
	
	\section{Results}
	\subsection{KNN Baseline\label{sec:KNN}}
	K-Nearest Neighbor (KNN) regression~\cite{bishop2006pattern} has been employed as a baseline. For the KNN regression, the data is temporally subsampled by a factor of 5. This is done to decrease the search time required for the KNN algorithm, without much information loss. The foot pressure data is sampled at 100Hz, i.e., a frame of foot pressure is recorded every 10 milliseconds. Since the data is sequential, picking every other frame of training data does not affect the temporal consistency of foot pressure data as the change in the heatmaps in 20 milliseconds is negligible. Pre-processing is carried out similar to training PressNET. The input pose data is normalized by mean and standard deviation of input, calculated using hip joint centered data by ignoring zero confidence values. The distance metric for the KNN algorithm is calculated as the mean of the distances between corresponding joints that do not have zero confidence values. This distance $d$ can be represented for two OpenPose human pose detections ($a$ and $b$) with confidence $c^a$ for $a$ and $c^b$ for $b$ and $J$ joints by: 
	\begin{equation}\label{eq1}
		d(a,b) = \frac{\sum_{j \in J}  \left\lVert \left(a_j - b_j\right)\right\rVert_2^2 *  \delta\!\left(c_j^a c_j^b>0\right)}{\sum_{j \in J} \delta\!\left(c_j^a c_j^b>0\right) + \epsilon }  
	\end{equation}
	where $\delta$ is the Kronecker Delta which is 1 if the argument is true and 0 if false, and a small $\epsilon$ avoids division by 0.  This enables KNN to adapt to missing joint detections in the human pose predictions. The KNN algorithm with K=1 is applied to all the six leave-one-subject-out splits. For each pose in the test split, which consists of data from any one subject, the corresponding ``nearest'' pose is picked from the training split consisting of data from the other five subjects.  The foot pressure map corresponding to this nearest neighbor is the prediction for the input pose in the training split. KNN with K=1 is a natural choice as a baseline because of the reasonable intuition/assumption that similar poses may lead to similar foot pressures. In our Leave-One-Out setting, KNN provides a measure of similarity between two poses of different subjects, thus establishing an upper-bound on foot pressure errors inherent in the dataset. 
	
	\begin{figure}[!t]
		\vspace{-1em}
		\begin{center}
			\includegraphics[width=.9\linewidth]{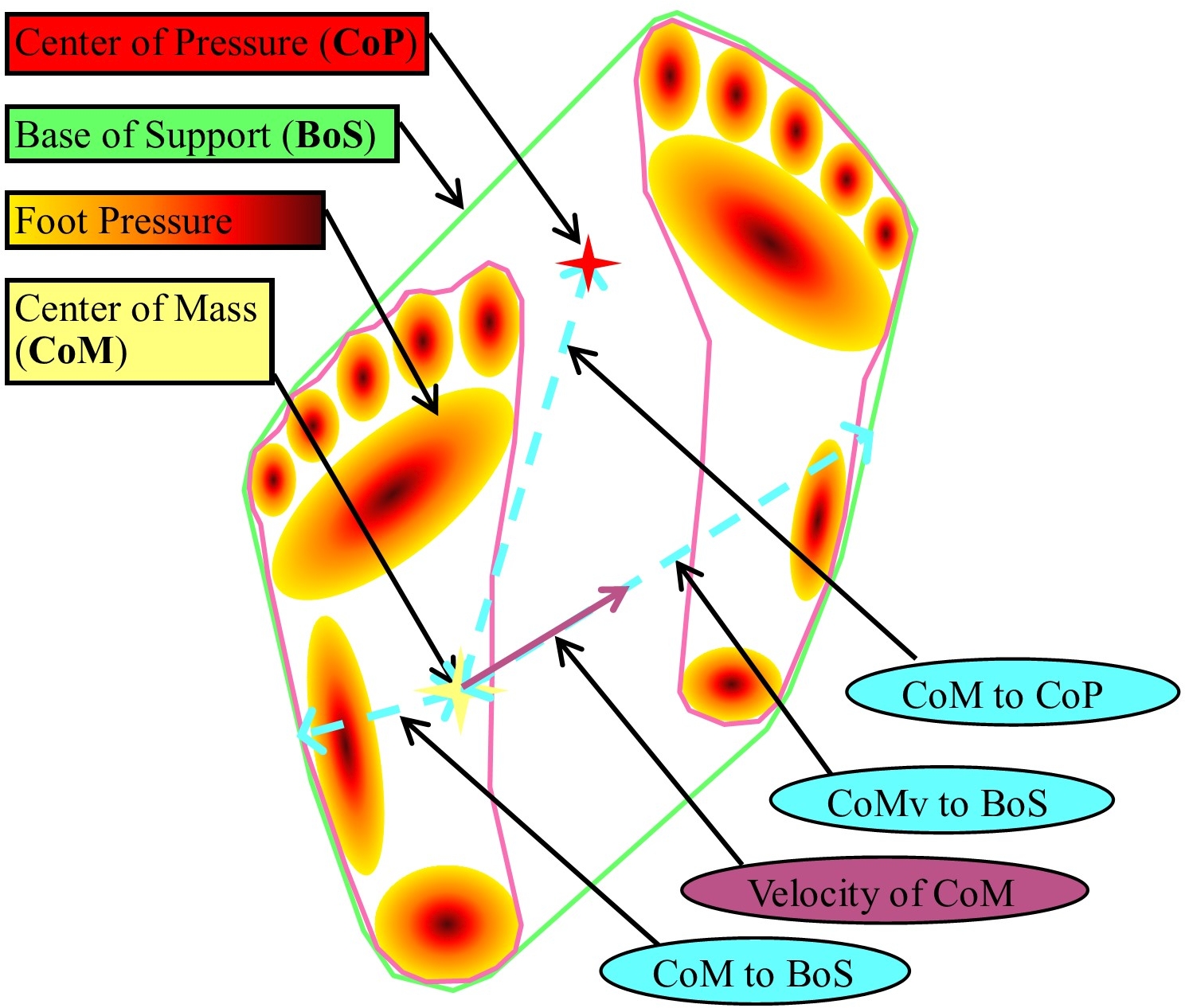}
		\end{center} \vspace{-1em}
		\caption[Overview of Stability and CoP]{Overview of concepts in stability. Depiction of the components of stability CoP, CoM, and BoS as well as the relation of those components to stability metrics.} \vspace{-15pt}
		\label{fig:sup:6} 
	\end{figure}

	\subsection{Stability}
	A major motivation for computing foot pressure maps from video is the application to stability analysis. Fundamental concepts used in stability analysis are illustrated in Figure~\ref{fig:sup:6}. These include Center of Mass (CoM), Base of Support (BoS), and Center of Pressure (CoP). CoM, also known as Center of Gravity, is the ground projection of the body's 3D center of mass~\cite{gos}. Generally speaking, human gait is stable if the CoM is contained withing the convex hull of the BoS, also called the support polygon~\cite{mrozowski2007analysis}.  If the CoM point is outside the support polygon, it is equivalent to the presence of an uncompensated moment acting on the foot, causing rotation around a point on the polygon boundary, resulting in instability and a potential fall. Center of Pressure (CoP), also known as the Zero Moment Point, is a point where the total moment generated due to gravity and inertia equals zero. Figure~\ref{fig:sup:6} shows a diagram of foot pressure annotated with the CoP, shown as a red star, with pressure from both feet shown as regions color-coded from low pressure (yellow) to moderate pressure (red) to high pressure (brown). Considering CoP as the ground reaction force and CoM as the opposing force, larger distances between the two 2D points could indicate reduced stability. Specifically, the CoP location relative to the whole body center of mass has been identified as a determinant of stability in a variety of tasks  \cite{hof2007equations,hof2008extrapolated,pai2003movement}. Note that the CoP is usually measured directly by force plates or insole foot pressure sensors, whereas in this paper we infer it from video alone. We quantitatively evaluate our results using ground truth data collected by insole foot pressure sensors.
	
	\subsection{Quantitative Evaluation}
	Two metrics for quantitative evaluation of our networks have been used:
	\begin{enumerate}[noitemsep,topsep=0pt]
		\item  Mean Absolute Error of Estimated Foot Pressure maps (kPa) as compared to ground truth pressure and
		\item  Euclidean ($\ell_2$) distance of Center of Pressure (mm) as compared to CoP calculated directly from ground truth foot pressure.
	\end{enumerate}
	To quantify these results with respect to physical units, the foot pressure data is un-normalized by reversing the pre-processing normalization.   
	
	\subsubsection{Mean Absolute Error of Predicted Foot Pressure}
	
	\begin{figure*}[!t] \centering
		\resizebox*{0.8\textwidth}{!}{
			\begin{tabular}{cc}
				\includegraphics[width=\linewidth]{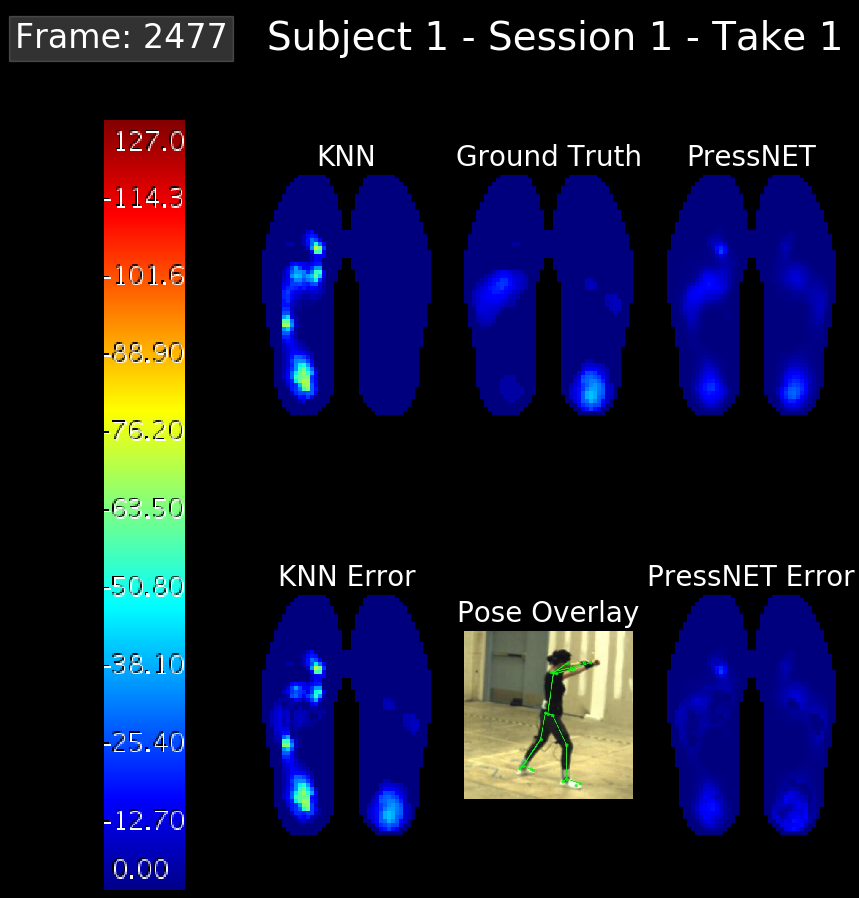}  & \includegraphics[width=1.025\linewidth]{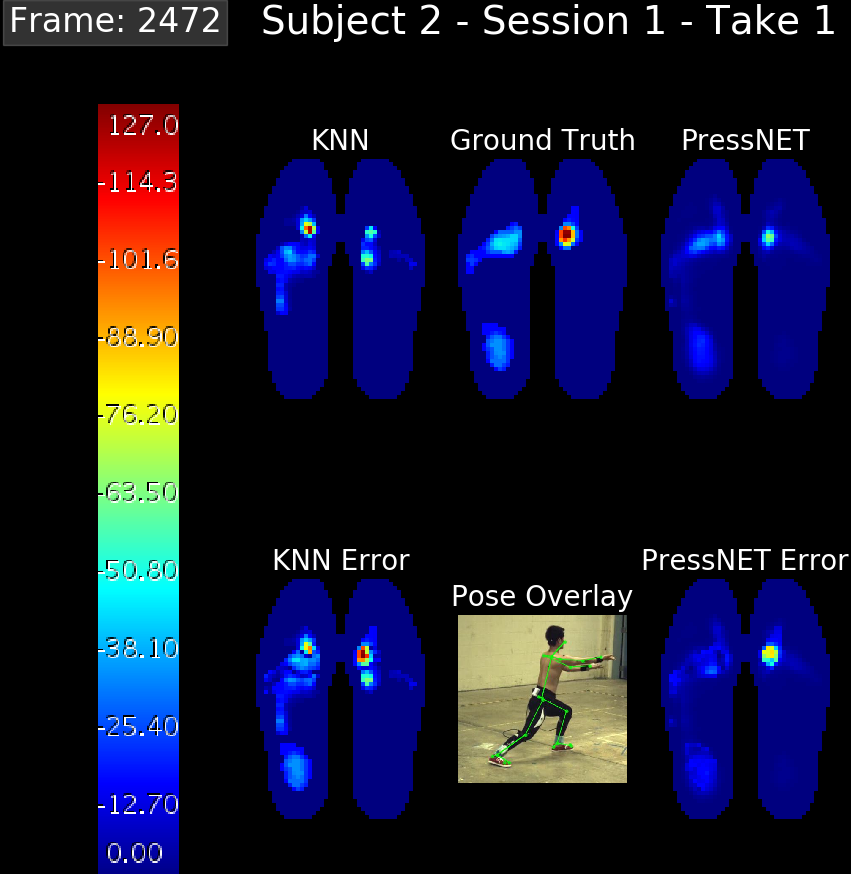}  \\
				\includegraphics[width=\linewidth]{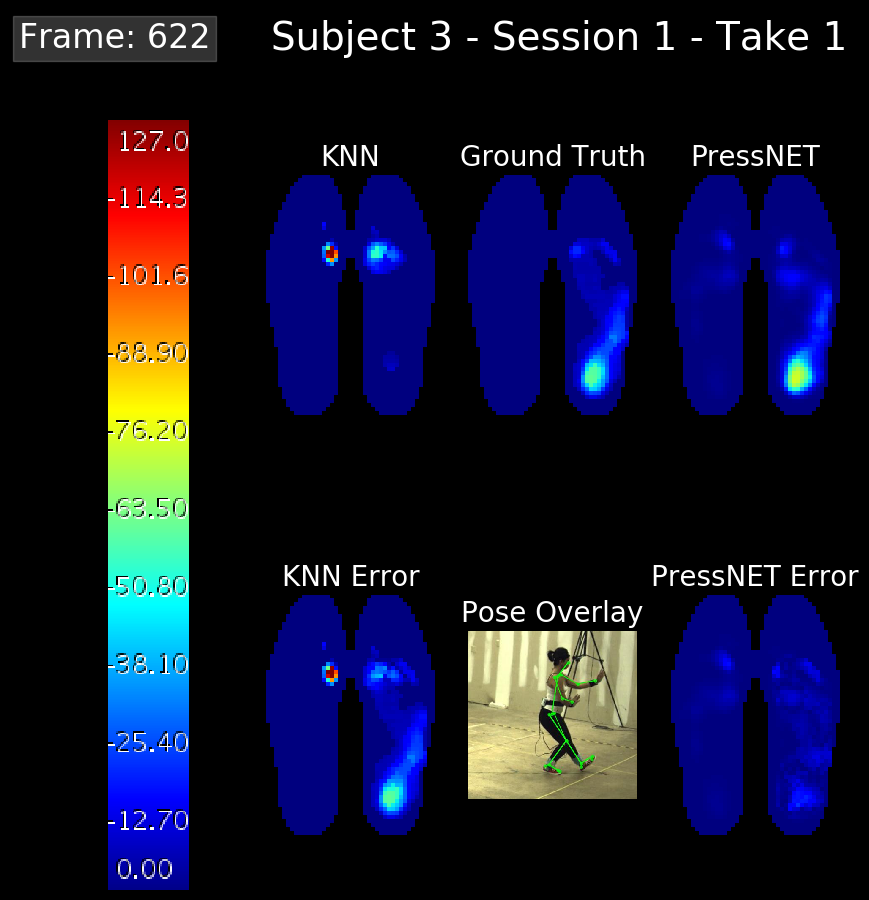}  & \includegraphics[width=\linewidth]{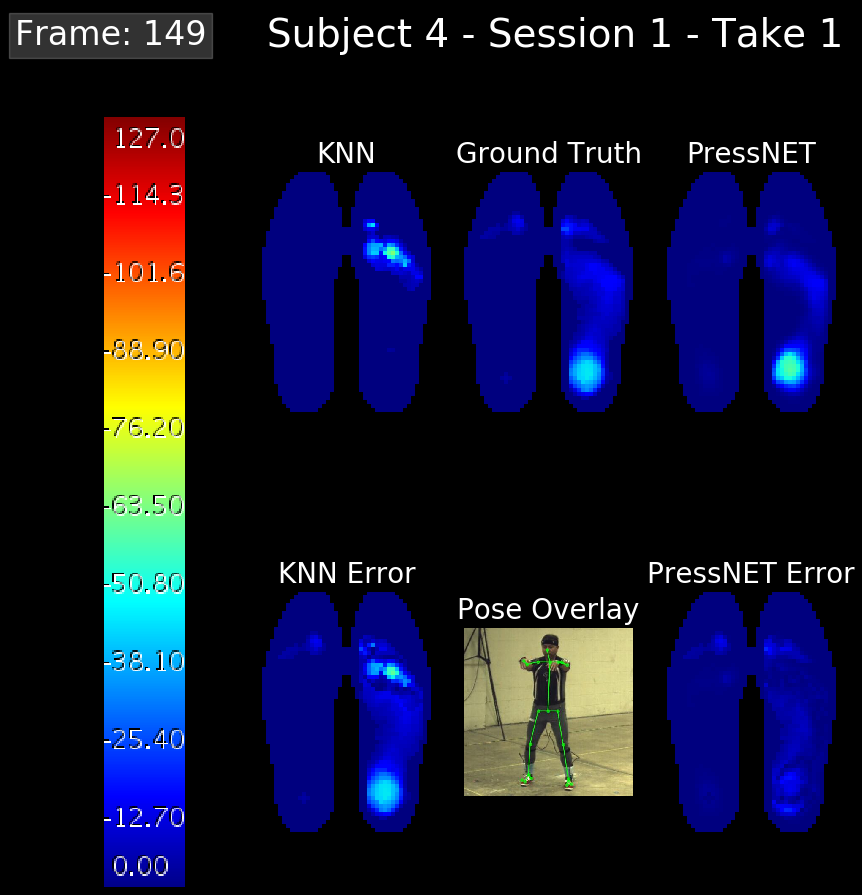} \\
				\includegraphics[width=\linewidth]{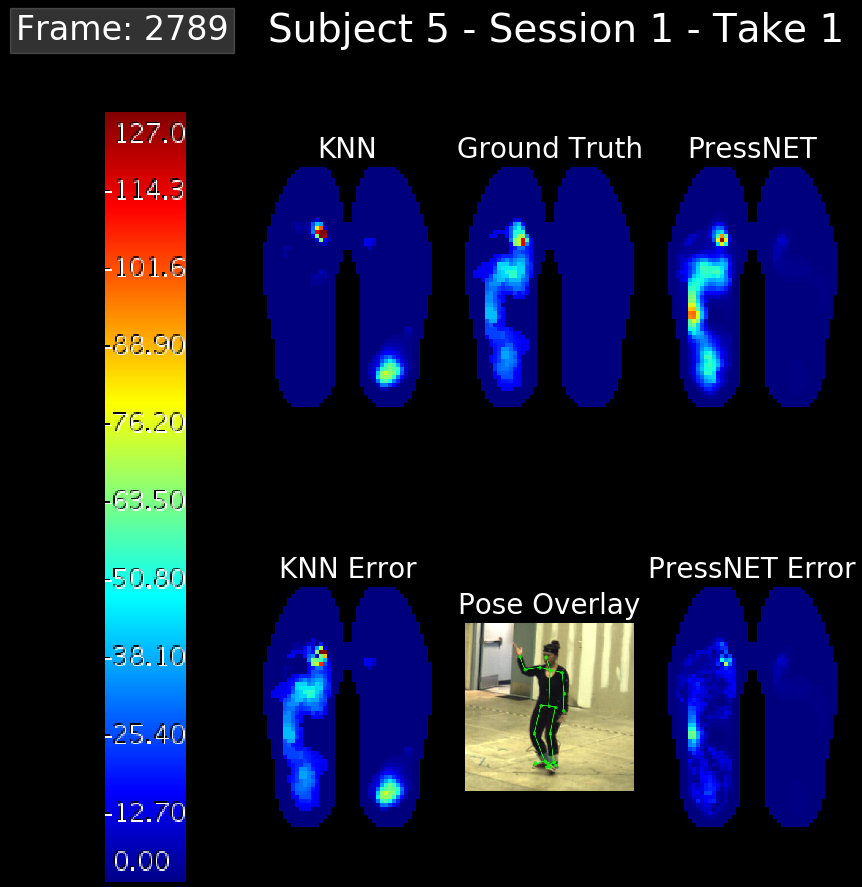} & \includegraphics[width=\linewidth]{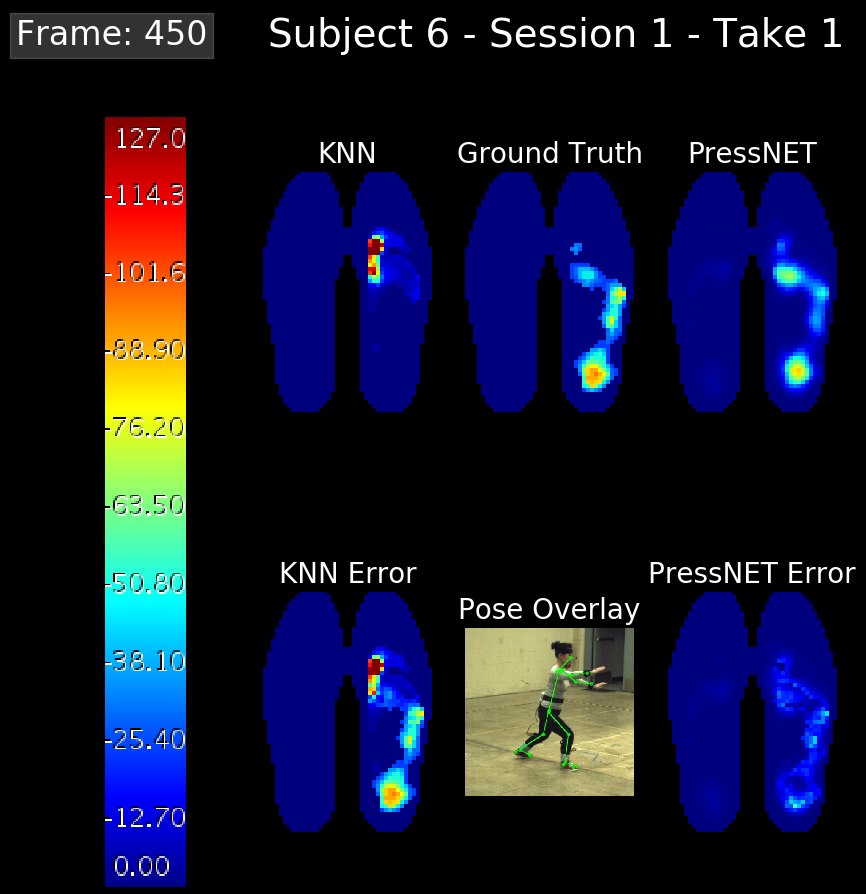} \\
			\end{tabular}
		}
		\caption{Example frames from each of the 6 subjects. Each frame provides KNN, Ground Truth, PressNET, KNN error from Ground Truth, PressNET error from Ground Truth, and the pose data extracted from the video frame.}
		\label{fig:screenshots}
	\end{figure*}
	\begin{figure*}[!t]
		\begin{center}
			\includegraphics[width=1\linewidth]{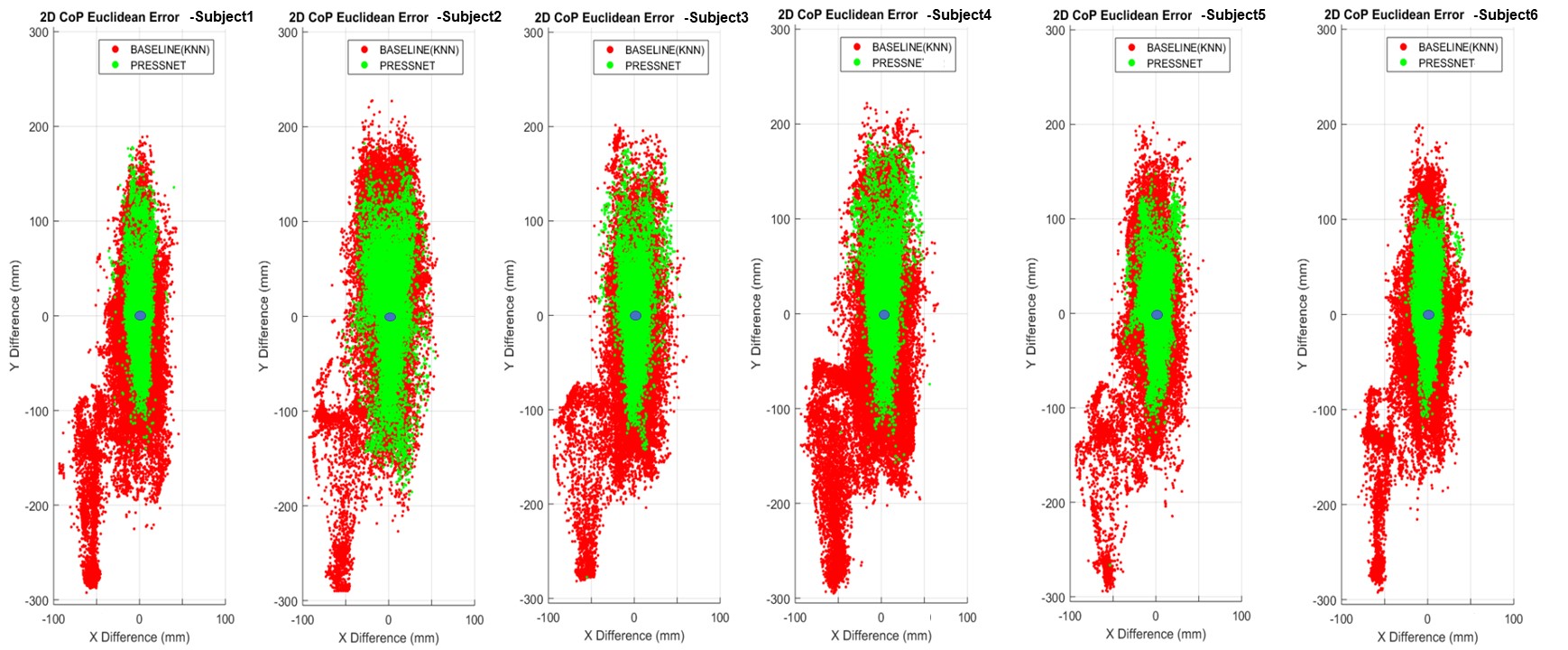}
		\end{center}\vspace{-10pt}
		\caption[CoP Euclidean error scatter Plots]{The 2D center of pressure error plots for KNN (red) and PressNET (green) difference from ground truth (blue) for each subject.}
		\label{fig:6} \vspace{-10pt}
	\end{figure*} 
	
	\begin{table}[!t] \centering
		\resizebox{\linewidth}{!}{
			\begin{tabular}{*{11}{c}} 
				
				& \multicolumn{10}{c}{KNN versus \textbf{PressNET} Testing Mean Absolute Errors (kPa)}  \\ \hline
				Subject & \multicolumn{2}{c}{Mean} & \multicolumn{2}{c}{Std} & \multicolumn{2}{c}{Median} & \multicolumn{2}{c}{Max} & \multicolumn{2}{c}{Min} \\ \hline 
				1 & 10.5 & \textbf{3.6} & 1.8 & \textbf{1.8} & 10.3 & \textbf{3.3} & 18.4 & \textbf{11.6} & 2.3 & \textbf{1.3}\\ \hline
				2 & 12.1 & \textbf{5.9} & 2.7 & \textbf{1.8} & 11.8 & \textbf{5.8} & 24.1 & \textbf{16.3} & 1.8 & \textbf{2.0}\\ \hline
				3 & 11.4 & \textbf{5.0} & 2.4 & \textbf{1.6} & 10.7 & \textbf{4.8} & 21.6 & \textbf{14.9} & 1.6 & \textbf{1.9}\\ \hline
				4 & 11.6 & \textbf{5.5} & 2.9 & \textbf{1.9} & 11.3 & \textbf{5.2} & 25.7 & \textbf{18.0} & 2.2 & \textbf{1.4} \\ \hline
				5 & 12.6 & \textbf{6.1} & 2.7 & \textbf{1.8} & 12.2 & \textbf{5.9} & 22.9 & \textbf{19.8} & 2.4 & \textbf{2.4}\\ \hline
				6 & 11.2 & \textbf{4.9} & 2.3 & \textbf{1.5} & 10.9 & \textbf{4.6} & 20.2 & \textbf{14.0} & 2.8 & \textbf{2.3}\\  \hline
				\rowcolor{Gray}
				Means & 11.6 & \textbf{5.16} & 2.47 & \textbf{1.73} & 11.2 & \textbf{4.93} & 22.15 & \textbf{15.76} & 2.18 & \textbf{1.88} \\ \toprule
			\end{tabular}
		} 
		\vspace{0pt} 
		\caption[Table of Mean Absolute Errors]{Results for KNN baseline and PressNET, showing the mean absolute error for each split of the data and the mean among the splits.  For every statistic, PressNET has lower  error than the baseline (lowest mean error values are shown in bold). }
		\label{tab:errortable} \vspace{-10pt} 
	\end{table}
	Mean absolute error $ E $ is used to quantify the difference between ground truth foot pressure $ Y $ and predicted foot pressure $\hat{Y} $ on $ N $ foot pressure prexels as:
	\begin{equation}
		E = \frac{1}{|N|} \sum_{n \in N} \left|Y_n - \hat{Y_n}\right|
	\end{equation}
	Mean across all the cross-validation splits is taken for our final accuracy rates.

	\par Table~\ref{tab:errortable} shows the mean absolute errors of predicted foot pressure for each data split, for both PressNET and the KNN baseline.  For PressNET, the worst two individuals to predict foot pressure on are Subjects 2 and 5.  This is also true for the nearest neighbor classifier as well, which could be because these subject foot pressures have a higher mean foot pressure than the other subjects and thus the networks are under-predicting the error.  PressNET has a mean classification error of 5.16 kPa which is less than double the 3 kPa measurement noise level of the foot pressure recording devices, as mentioned in Section~\ref{fp}.
	
	In order to test whether these results are significant, a two tailed paired t-test was performed between the mean errors for each frame for PressNET compared with KNN over each split of the data. This test showed that  the results from PressNET are statistically significantly better than the KNN baseline.

	\subsubsection{Computation for Center of Pressure}
	As a step towards analyzing gait stability from video, Center of Pressure (CoP) from regressed foot pressure maps of KNN and PressNET have been computed and quantitatively compared to ground truth (GT) CoP locations computed from the insole foot pressure maps. CoP is calculated as the weighted mean of the pressure elements (prexels) in the XY ground plane. The $\ell_2$ distance is used as a 1D metric to quantify the 2D error between ground truth and predicted CoP locations. Table~\ref{tab:CoP} shows the mean and standard deviations of $\ell_2$ errors calculated for the KNN baseline and PressNET. This table clearly shows that the mean distance error of COP computed from the PressNET network pressure map predictions is approximately 5-6 times smaller than that of KNN, with a standard deviation that is approximately 2-3 times smaller. The PressNET average error for all leave one subject out experiments yields a CoP Euclidean error of 10.52 $\pm$ 23.05 mm for the left foot and 9.63 $\pm$ 20.0 for the right.
	
	As shown in Figure~\ref{fig:6}, the distribution of PressNET distance errors are concentrated more tightly around zero millimeters, showing that the spatial accuracy of PressNET based CoP calculations are better than KNN, with smaller variance. As a point of reference, it is known  that a Center of Mass (CoM) localization accuracy of about 18 mm (or 1\% of the subject's height) is as accurate as the variation between multiple existing CoM calculation methods~\cite{virmavirta2014determining}.
	
	
	\begin{table}[!t] \centering 
		
		\resizebox{\linewidth}{!}{
			\begin{tabular}{|r||r|r|r|r||r|r|r|r|} \hline
				& \multicolumn{8}{|c|}{KNN (Baseline) versus \textbf{PressNET}} \\ \hline              
				& \multicolumn{4}{|c||}{Left Foot} & \multicolumn{4}{c|}{Right Foot} \\ \hline 
				
				
				Subject & \multicolumn{2}{|c|}{mean} & \multicolumn{2}{|c||}{std} & \multicolumn{2}{|c|}{mean} & \multicolumn{2}{|c|}{std} \\ \hline
				1       & 62.99 &  \textbf{7.52} & 64.75 & \textbf{18.37} & 55.59 &  \textbf{8.02} & 53.65 & \textbf{17.07} \\
				2       & 68.86 & \textbf{14.02} & 58.24 & \textbf{29.01} & 69.89 & \textbf{13.78} & 50.43 & \textbf{28.22} \\
				3       & 56.78 & \textbf{11.23} & 54.64 & \textbf{24.44} & 64.85 & \textbf{10.98} & 55.55 & \textbf{22.47} \\
				4       & 74.50 & \textbf{11.87} & 64.99 & \textbf{27.29} & 67.56 &  \textbf{9.40} & 55.63 & \textbf{20.20} \\
				5       & 47.63 & \textbf{10.14} & 47.23 & \textbf{21.33} & 51.21 &  \textbf{8.82} & 48.43 & \textbf{17.61} \\
				6       & 52.77 &  \textbf{8.34} & 55.92 & \textbf{17.84} & 42.45 &  \textbf{6.76} & 38.02 & \textbf{14.44} \\ \hline \rowcolor{Gray}
				All     & 60.59 & \textbf{10.52} & 57.63 & \textbf{23.05} & 58.59 &  \textbf{9.63} & 50.28 & \textbf{20.00} \\ \hline
			\end{tabular}
		}
		\vspace{1pt} 
		\caption{Mean and standard deviation of Euclidean errors in mm of CoP locations computed from predicted foot pressure maps of KNN and PressNET as compared to ground truth CoP evaluated on the left and right foot separately. \textbf{Bold} indicates minimum error.  See Figure~\ref{fig:6} for a display in 2D.}
		\label{tab:CoP} \vspace{-10pt} 
	\end{table}

	\subsection{Qualitative Evaluation}
	\subsubsection{Qualitative Evaluation of Mean Absolute Error}
	Figure~\ref{fig:screenshots} visualizes foot pressure predictions and their mean absolute errors for some example frames.  The foot pressure predictions, ground truth and absolute difference heatmaps are rendered on the same color scale. The color bar in each sub-frame represents foot pressure intensity in kilopascals (kPa). It starts from a shade of blue, representing 0 foot pressure to a dark shade of red, corresponding to maximum foot pressure observed during the performance. The colors in between represent different levels of pressure intensities between 0 and the maximum. 
	It can be visually observed that the PressNET errors are small compared to KNN errors. 
	In addition to the qualitative comparison by visualization, the respective mean absolute errors with respect to ground truth frames have also been calculated to provide a quantitative comparison of performance. The frames have been chosen to show the ability of PressNET to generalize to different poses, similar poses from different subjects and different views with respect to a fixed camera. The frames have also been chosen to show some failure cases.
	
	It is evident that the heatmaps generated by PressNET are more similar to ground truth heatmaps. This is supported by the mean absolute frame errors of the networks. KNN results are visually very poor when compared to  PressNET because 1-KNN is merely picking the frame with the shortest distance between joints in a cross-subject evaluation. As the style of performance and body dynamics differs for each subject, KNN is unable to generalize to a change in subjects, leading to high mean absolute error.

	Observing foot pressure predictions temporally over a sequence of frames, it was observed that KNN predictions are highly inconsistent and fluctuating, whereas the PressNET predictions are temporally smooth and consistent. Since the system operates on a per-frame basis, KNN picks the frame with the nearest pose in the dataset to the current frame. This makes the predictions fluctuate over time. Even though our network is trained using the same per-frame basis mechanism, it has learned to predict the mean foot pressure heatmaps over a window of frames, i.e., our network has learned to be temporally stable, making the predictions smooth and more similar to ground truth. 
	
	\subsubsection{Qualitative Evaluation of Center of Pressure}
	Table~\ref{tab:CoP} showed mean and standard deviation of Euclidean errors in millimeters of CoP locations computed from predicted foot pressure maps of KNN and PressNET as compared to ground truth CoP evaluated on the left and right foot separately. Figure~\ref{fig:6} shows the 2D CoP distance error scatter plots for Subjects 1, 2, 3, 4, 5 and 6. It can be observed from the figures that the spread of errors of CoP for PressNET is significantly lower that that of KNN and concentrated around (0,0). 
	
	\section{Summary and Conclusion}
	In this research, the feasibility of regressing foot pressure from 2D joints detected in video has been explored. This is the first work in the computer vision community to establish a direct mapping from 2D human body kinematics to foot pressure dynamics. The effectiveness of our PressNET network has been shown both quantitatively and qualitatively on a challenging, long, multi-modality Taiji performance dataset. Statistically significant improved results  over a standard K-Nearest Neighbor method in foot pressure map estimation from video have been demonstrated. The outcome of PressNET is encouraging since it is also within the range of inter-subject variance of the same pose observed (Figure~\ref{data_stat}).
	Furthermore, we demonstrate the use of regressed foot pressure results for estimation of {\em Center of Pressure}, a key component of postural and gait stability. The errors (Table~\ref{tab:CoP}) are within the accepted range for kinesiology studies of Center of Mass (CoM) \cite{virmavirta2014determining}, a corresponding dynamics concept to CoP in stability analysis.  
	
	We hope to extend this work to include more aspects of human body dynamics such as regressing directly to muscle activations, weight distributions, balance, and force. Our goal is to build {\em precision computer vision} tools that estimate various human body dynamics using passive and inexpensive visual sensors, with outcomes validated using bio-mechanically derived data (rather than approximations by human labelers). We foresee  introducing a new and exciting sub-field in computer vision going beyond visually satisfactory human joint/pose detection to the more challenging problems of capturing accurate, quantifiable human body dynamics for scientific applications.  
	
	\section{Acknowledgments}
	We would like to thank the six volunteers who contributed 24-form Taiji performances to this study. We would like to acknowledge Andy Luo for his help in rendering the images and videos for demonstration. We thank the College of Engineering Dean's office of Penn State University for supporting our motion capture lab for research and education. This human subject research is approved through Penn State University IRB Study8085. This work is supported in part by NSF grant IIS-1218729.
	
	{\small
		\bibliographystyle{ieee}
		\bibliography{ms}
	}
	
\end{document}